\documentclass{article}

\usepackage[numbers, compress]{natbib}

\usepackage{bm}

 \usepackage[main, final]{neurips_2026}

\usepackage[utf8]{inputenc} 
\usepackage[T1]{fontenc}    
\usepackage{hyperref}       
\usepackage{url}            
\usepackage{booktabs}       
\usepackage{amsmath}        
\usepackage{amsfonts}       
\usepackage{nicefrac}       
\usepackage{microtype}      
\usepackage{xcolor}         
\usepackage{caption}        
\usepackage{multirow}
\usepackage{multicol}
\usepackage{enumitem}       
\usepackage{cleveref}
\usepackage{multirow}
\usepackage{graphicx}
\usepackage{subcaption}
\usepackage{amssymb}
\usepackage{array}
\usepackage[most]{tcolorbox}

\tcbuselibrary{listings, breakable}

\usepackage{wrapfig} 

\usepackage{amsmath,amsfonts,bm}
\usepackage{amsthm}
\usepackage{wrapfig}
\usepackage{tikz}
\usetikzlibrary{shapes, positioning, arrows}
\usetikzlibrary{shapes.geometric}
\usepackage{tabularx}
\usepackage{xspace}
\usepackage{multicol}
\usepackage{multirow}
\usepackage{bbm}
\usepackage{todonotes}
\usepackage{quiver}

\theoremstyle{plain}

\def\eqref#1{equation~\ref{#1}}

\def\1{\bm{1}}

\def\vq{{\bm{q}}}

\def\vy{{\bm{y}}}

\def\mX{{\bm{X}}}

\DeclareMathAlphabet{\mathsfit}{\encodingdefault}{\sfdefault}{m}{sl}
\SetMathAlphabet{\mathsfit}{bold}{\encodingdefault}{\sfdefault}{bx}{n}

\usepackage{amsmath,amsfonts,bm}

\def\eqref#1{equation~\ref{#1}}

\def\1{\bm{1}}

\def\vq{{\bm{q}}}

\def\vy{{\bm{y}}}

\def\mX{{\bm{X}}}

\DeclareMathAlphabet{\mathsfit}{\encodingdefault}{\sfdefault}{m}{sl}
\SetMathAlphabet{\mathsfit}{bold}{\encodingdefault}{\sfdefault}{bx}{n}

\DeclareMathOperator*{\argmax}{arg\,max}

\newcolumntype{C}[1]{>{\centering\arraybackslash}p{#1}}
\newcolumntype{R}[1]{>{\raggedleft\arraybackslash}p{#1}} 

\tcbset{
  promptbox/.style={
    breakable,
    colback=blue!5,
    colframe=blue!40,
    fontupper=\footnotesize\ttfamily,
    boxrule=0.4pt,
    arc=2pt,
    left=4pt, right=4pt, top=4pt, bottom=4pt,
    coltitle=white,
  }
}

\title{\textsc{RelAgent}: LLM Agents as Data Scientists for Relational Learning}

\author{
Xingyue Huang\textnormal{\textsuperscript{1}\thanks{A substantial part of this research was conducted while the author was a predoctoral fellow at AITHYRA.}\quad}
Louis Tichelman\textnormal{\textsuperscript{2,3}\quad}
Jinwoo Kim\textnormal{\textsuperscript{4}\quad}
Krzysztof Olejniczak\textnormal{\textsuperscript{1}\quad}\\
\textbf{{\.I}smail {\.I}lkan Ceylan}\textnormal{\textsuperscript{2,3,1}}\\
\textsuperscript{1}University of Oxford\quad
\textsuperscript{2}TU Wien\quad
\textsuperscript{3}AITHYRA\quad
\textsuperscript{4}KAIST
}

\begin{document}

\maketitle

\begin{abstract}
Relational learning is a challenging problem that has motivated a wide range of approaches, including graph-based models (e.g., graph neural networks, graph transformers), tabular methods (e.g., tabular foundation models), and sequence-based approaches (e.g., large language models), each with its own advantages and limitations. We propose \textsc{RelAgent}, an LLM-based autonomous data scientist for relational learning, which operates in two phases. 
In the \emph{search phase}, an LLM agent uses database, validation, and evaluation workspace tools to construct SQL feature programs and select a predictive model. In the \emph{inference phase}, the resulting program is executed without further LLM calls.
The final predictor consists of SQL queries and a classical model, enabling fast, deterministic, and intrinsically interpretable predictions: features are human-readable queries, and predictions depend only on the resulting query-defined feature map, enabling scalable deployment using standard database systems.
\end{abstract}

\section{Introduction}
\vspace{-0.1em}

Relational databases are the standard paradigm for storing, managing, and processing structured data~\citep{codd1970relational}. In practice, predictive modeling over such data is driven by feature construction: analysts write SQL queries that join tables along foreign-key paths, aggregate signals over time windows, and construct features such as counts, recency statistics, or activity trends for downstream models~\citep{relbench,relbenchv2,kanter2015deep}.
The space of possible features is vast, and identifying useful combinations often requires iterative trial-and-error over joins, filters, and aggregations. This process is time-consuming, requires substantial domain expertise, and becomes increasingly difficult to scale as database schemas grow in complexity.

Machine learning over relational databases, commonly referred to as relational learning, aims to automate this process. 
Existing approaches address this challenge through different representations, access patterns, and computational mechanisms (see \Cref{tab:comparison}).
\emph{Tabular methods}, including supervised tabular models~\citep{chen2016xgboost,lightgbm, autogluon} and tabular foundation models~\citep{tabpfnv2,xu2026rdblearn}, operate on flat feature matrices and can be highly effective, but applying them to relational databases requires either substantial feature engineering or generic aggregations that may fail to capture task-specific relational interactions. \emph{Graph-based models}, such as graph neural networks~\citep{graphsage,gat,relgt}, graph transformers~\citep{relgt}, and relational foundation models~\citep{ranjan2026rt,griffin,fey2025kumorfm-v1,hudovernik2026kumorfm}, typically operate on a graph induced by the database through entity-centric subgraphs. While powerful, these methods rely on a fixed mechanism for accessing relational structure, where the effective context, such as the neighborhood size and depth of aggregation, is chosen in advance and may not necessarily align with the downstream task. \emph{Sequence-based approaches}, such as large language models (LLMs)~\citep{wydmuch24,talklikeagraph} applied to serialized tables or graphs, offer flexible reasoning interfaces, but their inductive bias is not naturally aligned with structured relational prediction. These limitations point to a complementary direction: keeping relational structure external and executable, while automating the search for task-relevant queries.

\begin{figure}[t]
    \centering
    \vspace{-2em}
    \includegraphics[width=0.9\textwidth]{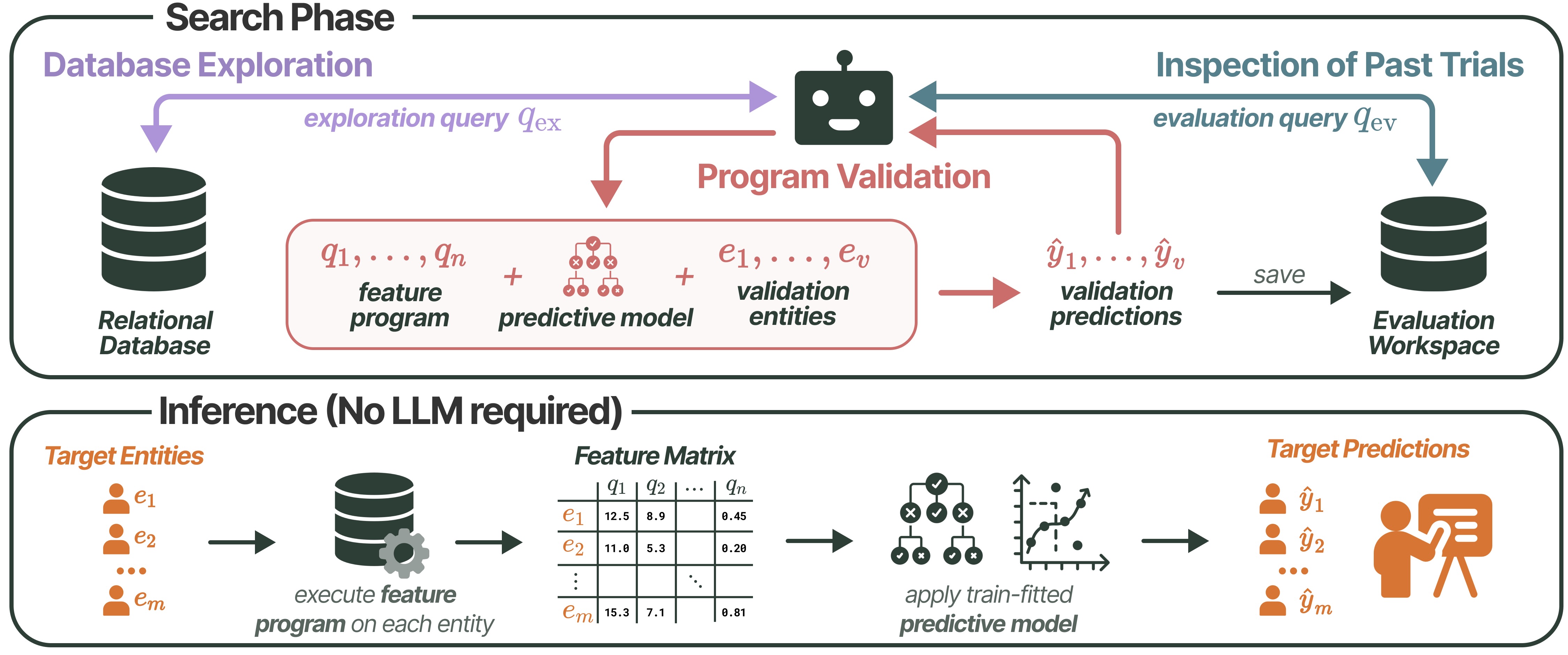}
    \caption{ 
    \textbf{\textsc{RelAgent}.} During the search phase, an LLM agent iteratively proposes and refines a \textit{feature program} consisting of SQL feature queries $\{q_1,\dots,q_n\}$ and a predictive model configuration $\varphi$ to solve a given task. The agent uses three tools: \textbf{(1) database exploration} via read-only SQL exploration queries, \textbf{(2) program validation} by executing candidate programs on a validation set and receiving performance metrics, and \textbf{(3) inspection of past trials} in the \textit{Evaluation Workspace} via evaluation queries. Once a final program is selected, the agent is no longer needed at inference time.
    }
    \label{fig:MethodOverview}
  \vspace{-1em}
\end{figure}

\textbf{Our approach.} We propose \textsc{RelAgent}, a framework that casts relational learning as an \emph{agentic search process} over query-based features and predictive models. 
As illustrated in Figure~\ref{fig:MethodOverview}, 
\textsc{RelAgent}  operates in two phases:
\begin{enumerate}[label=(\arabic*),leftmargin=20pt,itemsep=1pt, topsep=1pt]
    \item \emph{Search phase}: An LLM agent interacts with the system through three query interfaces: (i)~\textbf{exploration queries} that inspect the database, (ii) \textbf{feature queries} that construct predictive features, and (iii) \textbf{evaluation queries} that analyze predictions and diagnostics stored in an evaluation workspace. The agent iteratively proposes feature programs, selects a downstream predictive model (e.g., XGBoost~\citep{chen2016xgboost}, LightGBM~\citep{lightgbm}), and refines program and model choice based on this feedback. 
    \item \emph{Inference phase}: The selected feature program is executed on the target entities, and the resulting features are passed to the predictor fitted from the training examples.
\end{enumerate}
More broadly, \textsc{RelAgent} defines predictors over query-induced feature maps, with the hypothesis class parameterized by executable queries rather than fixed architectures. The final predictor consists only of an SQL feature program and a predictive model, yielding predictions that are fast, deterministic, and intrinsically interpretable: each prediction can be traced to the selected queries, their materialized feature values, and the downstream model’s feature usage.
This also gives a different view of relational inductive bias: two target examples with identical outputs under the selected SQL features receive identical predictions, so invariance is induced by the selected queries rather than by a fixed graph encoding, sequence ordering, or neural architecture.

\noindent\textbf{Contributions.}
We summarize our main contributions as follows:
\begin{enumerate}[label=$\bullet$,leftmargin=20pt,itemsep=1pt, topsep=1pt]
    \item We introduce \textsc{RelAgent}, an agentic framework that formulates relational learning as an autonomous search process over query-based features and predictive models using an LLM interacting with relational databases via tools.
    \item We propose an execution-grounded learning loop with validation feedback and an evaluation workspace, enabling iterative refinement through structured analysis of prior trials.
    \item We decouple search and inference, yielding a final predictor that does not require an LLM at inference time, and is fast, deterministic, and interpretable. 
    \item We formulate relational learning over query-defined feature maps, with invariance induced by executable queries rather than fixed architectural biases.
    \item We empirically evaluate \textsc{RelAgent} on relational benchmarks, demonstrating competitive performance with strong baselines while retaining interpretability and scalability.
\end{enumerate}

\section{Related work}
\vspace{-0.1em}

\noindent\textbf{Relational deep learning.}
Predictive modeling over relational databases traditionally relies on manually materializing task-specific feature tables via joins and aggregations, followed by supervised tabular models such as XGBoost~\citep{chen2016xgboost}, LightGBM~\citep{lightgbm}, or AutoGluon~\citep{autogluon} -- a workflow that requires substantial engineering and does not transfer across schemas or targets.
Relational Deep Learning (RDL) avoids manual flattening by treating databases as temporal heterogeneous graphs (rows as nodes, primary-foreign key links as edges, training tables as targets)~\citep{relbench}, underlying GNN- and Graph Transformer-based models such as RelGNN~\citep{relgnn} and RelGT~\citep{relgt}. More recently, relational foundation models, including Griffin~\citep{griffin}, Relational Transformers~\citep{ranjan2026rt}, KumoRFM-v1~\citep{fey2025kumorfm-v1}, and KumoRFM-v2~\citep{hudovernik2026kumorfm}, aim to achieve stronger transfer across schemas, databases, and tasks. Another line of work directly utilizes tabular or language foundation models, with examples including GNN+TabPFN-2.5~\citep{gnntabpfn}, RDBLearn~\citep{xu2026rdblearn}, 
and LLM-based predictors that serialize relational database as context~\citep{ranjan2026rt,wydmuch24,relllm}. 
RDL is also related to the broader effort of building graph and knowledge graph foundation models~\citep{ultra,motif,hyper,trix,flock}, but is specialized to relational databases with schemas and temporal prediction targets.

\begin{table}[t]
\centering
\footnotesize
\caption{Comparison of relational learning approaches.}
\label{tab:comparison}
\setlength{\tabcolsep}{3.8pt}
\begin{tabular}{l p{2cm} p{2.25cm} p{2.3cm} p{2.2cm} p{2cm}}
\toprule
\textbf{Method} & \textbf{Representation} & \textbf{Access} & \textbf{Mechanism} & \textbf{Inductive bias} & \textbf{Interpretability} \\
\midrule

Tabular 
& Tables 
& Internalized (flatten) 
& Trees / Attention 
& Row perm.-equiv. 
& Feature-level \\

GNNs 
& Graph 
& Internalized (local) 
& Message passing 
& Perm.-equiv. 
& Post-hoc \\

Graph Transf.
& Graph (with PE)
& Internalized (global) 
& Attention 
& Perm.-equiv. 
& Post-hoc \\

LLMs 
& Sequence 
& Internalized (text)
& Attention 
& Order-sensitive 
& Post-hoc \\

\midrule
\textsc{RelAgent} 
& Database 
& External (queries) 
& SQL + Predictors
& Query-defined
& Intrinsic \\
\bottomrule
\end{tabular}
\end{table}

\noindent\textbf{LLMs for structured data.}
LLMs can be applied to structured data by serializing tables and schemas into text, enabling tasks such as tabular prediction~\citep{tabert} and 
text-to-SQL~\citep{tapex}. Row-level serialization can support few-shot tabular learning, but performance remains sensitive to prompt design and representation choices~\citep{tabllm,dinsql}.
Similarly, graph-structured data can be encoded as text, though such approaches are often brittle to formatting and neighborhood choices~\citep{talklikeagraph}. Later methods add structure-aware adaptations, including graph prompting and instruction tuning~\citep{graphgpt,instructglm,graphicl}, or inject structure via graph tokenization~\citep{graphtoken,llaga}.
More recently, tool use enables LLMs to interact with structured data through explicit operations rather than pure prompting. ReAct~\citep{react} interleaves reasoning with tool calls, Toolformer~\citep{toolformer} learns API use, and extensions such as Plan-and-Execute~\citep{wang2023plan} and Reflexion~\citep{reflexion} add planning and self-correction. Graph-specific variants further enable structured reasoning via tool invocation~\citep{graphtoolinstruction} or executable programs~\citep{finkelshteinactions}. Unlike approaches that internalize relational structure by serializing tables or graph neighborhoods into the LLM context, \textsc{RelAgent} keeps the database external and accessible through SQL.

\noindent\textbf{LLMs as feature engineers.}
Beyond using LLMs as direct predictors, recent work treats them as feature engineers that generate task-specific transformations for downstream models. Early approaches largely focus on single-table settings: CAAFE~\citep{caafe} and FeatLLM~\citep{featllm} show that LLMs can synthesize semantically meaningful features or rule-based transformations for tabular data, improving few-shot performance while avoiding LLM usage at inference time. More recent systems cast feature engineering as an iterative search process: LLM-FE~\citep{abhyankar2025llmfe} formulates feature construction as program search guided by validation feedback, and OCTree~\citep{octree} further incorporates decision tree reasoning to iteratively refine individual rules.
For multi-relational databases, ReFuGe~\citep{kim2026refuge} extends LLM-based feature generation through a fixed sequential pipeline with aggregate validation feedback and a fixed downstream model. \textsc{RelAgent} instead exposes three query interfaces: exploration queries over the database, feature queries used as executable programs, and evaluation queries over a persistent workspace of row-level predictions and trial histories. This enables targeted self-correction and joint search over SQL features and downstream models, while keeping the deployed predictor intrinsically interpretable.

\section{Preliminaries}
\vspace{-0.1em}
\label{sec:preliminaries}
\noindent\textbf{Relational databases.}
A \textit{relational database} is a collection of tables
\(\mathcal{D}=\{\mathcal{T}_1,\dots,\mathcal{T}_k\}\).
Each table contains rows representing entities or events, with typed columns such as numerical or categorical attributes, timestamps, and primary/foreign keys.
We denote the schema by \(\mathcal{S}(\mathcal{D})\), which includes table names, column names, types, and primary-foreign key relations defining links across tables.

\noindent\textbf{Entity prediction tasks over relational databases.}
A supervised entity-level prediction task consists of a relational database (called \emph{context}) \(\mathcal D\) and target tables
\(\mathcal{T}_{\mathrm{train}}, \mathcal{T}_{\mathrm{val}}, \mathcal{T}_{\mathrm{test}}\).
Each target row is of the form \((e,t,y)\), where \(e\) is the target entity, \(t\) is the prediction timestamp, and \(y\in\mathcal Y\) is a classification or regression label. The target tables are split chronologically by cutoffs
\(t_{\mathrm{train}}\!<t_{\mathrm{val}}\!<t_{\mathrm{test}}\), and \(\mathcal D\) contains only records available up to the test cutoff.

We enforce \emph{point-in-time correctness}: for each target row \((e,t,y)\), feature construction may use only source records with a timestamp strictly before \(t\). For validation and test predictions, we augment the context with training labels,
\(
\widetilde{\mathcal D}=\mathcal D\cup\{\mathcal T_{\mathrm{train}}\},
\)
and learn a predictor
\(
g:(\widetilde{\mathcal D},e,t)\mapsto \hat y.
\)
Thus, validation and test predictions may use historical relational records and training labels, but never validation or test labels.

\noindent\textbf{SQL queries.} We distinguish three types of queries. Let \(\mathcal{L}\) denote the class of SQL queries available to the agent. An \textit{exploration query} $q_e \in \mathcal{L}$ is a read-only query over the context database $\mathcal{D}$ used for schema and data inspection. A \textit{feature query} $q \in \mathcal{L}$ maps a target entity $e$, timestamp $t$, and context $\widetilde{\mathcal D}$ to one or more feature values, $q(\widetilde{\mathcal D},e,t) \in \mathcal{X}^{d_q}$, using only records before \(t\); a collection of feature queries $\vq=(q_1,...,q_n)$ is a \textit{feature program}. An \textit{evaluation query} is a read-only query over the evaluation workspace, used only for search-time diagnostics and never as a predictive feature.

\noindent\textbf{Feature-construction map.}
For a target table \(\mathcal{T}\) with \(m\) rows, a feature program \(\vq=(q_1,\dots,q_n)\) materializes and aligns query outputs into
\(
\mX=\Phi_{\vq}(\widetilde{\mathcal D},\mathcal{T})\in\mathcal{X}^{m\times d},
\)
where \(d=\sum_{j=1}^n d_{q_j}\).

\section{Methodology}
\vspace{-0.1em}

We present \textsc{RelAgent}, an agentic framework that uses an LLM agent to autonomously search over SQL-derived features and predictive models (such as XGBoost~\citep{chen2016xgboost}, LightGBM~\citep{lightgbm}, etc.) for relational learning via tool-use.
Given an augmented database context \(\widetilde{\mathcal{D}}\) and a validation target table \(\mathcal{T}_\text{val}\), the framework has two stages:
\begin{enumerate}[noitemsep,nolistsep,label=(\arabic*),leftmargin=*]
    \item \textbf{Search phase.}
    The agent iteratively explores the database, validates candidate feature programs and predictive models on $\mathcal{T}_\text{val}$, and inspects previous trials to select a feature program and model.
    \item \textbf{Inference phase.}
    The best validated program and model are deployed without further LLM calls: the program produces feature vectors, and the fitted predictive model outputs predictions.
\end{enumerate}

\subsection{Search phase}
\vspace{-0.1em}

Let \(\mathcal{L}\) be the class of SQL queries available to the agent and \(\mathcal{P}\) the class of predictive models, where each \(\varphi\in\mathcal{P}\) determines a model family and its hyperparameters. We write \(\mathcal{L}^{\ast}=\bigcup_{n\geq 1}\mathcal{L}^n\) for the set of finite-length feature programs. The agent searches over the joint space of feature programs and predictive models, $\mathcal{L}^{\ast}\times\mathcal{P}$, to identify a candidate \((\vq,\varphi)\) with strong validation performance. We formalize the search as a stochastic policy over a structured action space \(\mathcal{A}\).

Let $a_t \in \mathcal{A}$ denote the action taken at step $t$ and \(o_t\) the tool observation returned after executing $a_t$. Given the interaction history $H_t = \bigl(\widetilde{\mathcal{D}},\, \mathcal{T}_{\mathrm{val}},\, (a_1,o_1),\dots,(a_{t-1},o_{t-1})\bigr)$ at step $t$, the agent samples its next action as
\(
a_t \;\sim\; \pi(\cdot \mid H_t),
\)
where the policy $\pi$ is implemented by an LLM. The action space can be divided into three action types:

\noindent\textbf{Database exploration.}
The agent can inspect the augmented database \(\widetilde{\mathcal D}\) and the label-masked validation table \(\mathcal T_{\mathrm{val}}\) using two read-only tools:
\begin{enumerate}[label=(\arabic*),itemsep=0pt, topsep=0pt, parsep=0pt, partopsep=0pt]
    \item \texttt{execute\_query()}: runs read-only SQL exploration queries;
    \item \texttt{get\_table\_info()}: returns table schemas \(\mathcal{S}(\widetilde{\mathcal D})\), primary/foreign keys, and row counts.
\end{enumerate}

 This allows the agent to identify relevant tables, columns, timestamp fields, primary-foreign key paths, and candidate temporal aggregations. For instance, the agent typically first inspects table schemas and then issues aggregate queries to summarize entity frequencies, timestamp ranges, missingness patterns, and relationships between tables. These exploratory queries provide context for subsequent feature construction, but need not directly be used as predictive features unless the agent later includes them in a candidate program.

\noindent\textbf{Program validation.} 
The agent can submit a tuple $(\vq, \varphi)$ consisting of a feature program $\vq$ and a predictive model $\varphi$ for validation through the tool \texttt{validate\_program()}. 
The model $\varphi \in \mathcal{P}$ specifies both the model family and its associated hyperparameters. For regression tasks, it additionally includes optimization-related choices, such as the type of regularization (e.g., $L_1$, $L_2$), the loss function (e.g., Huber loss), and potential input transformations (e.g., a log-target transformation may be applied, whereby the model is trained on $\log(1 + y)$). 
The validation tool first produces features for training and evaluation entities via the feature construction map induced by $\vq$.
\[
\mX_{\mathrm{train}}
=
\Phi_{\vq}(\widetilde{\mathcal D},\mathcal T_{\mathrm{train}}),
\qquad
\mX_{\mathrm{val}}
=
\Phi_{\vq}(\widetilde{\mathcal D},\mathcal T_{\mathrm{val}}).
\]
Let \(\vy_{\mathrm{train}}\) and \(\vy_{\mathrm{val}}\) denote the label vectors extracted from
\(\mathcal T_{\mathrm{train}}\) and \(\mathcal T_{\mathrm{val}}\), respectively.
A~predictor $f$, initialized w.r.t the submitted model configuration \(\varphi\), is fitted to the training data \((\mX_{\mathrm{train}},\vy_{\mathrm{train}})\).
Consequently, the tool computes 
\[
\hat{\vy}_{\mathrm{val}}
=
f\!\left(\mX_{\mathrm{val}}\right),
\qquad
\mathrm{score}_{\mathrm{val}}(\vq,\varphi)
=
M\!\left(
\hat{\vy}_{\mathrm{val}},
\vy_{\mathrm{val}}
\right),
\]
where \(M\) is the task-specific validation metric, e.g., AUROC for binary classification or negative MAE for regression. 
The validation response also returns execution feedback, including SQL errors, timeout failures, and model-training errors, allowing the agent to revise later candidate programs.

\noindent\textbf{Reflection via evaluation workspace.} 
After each validation trial, the validation tool records the candidate program and validation artifacts in an \textit{evaluation workspace}. The evaluation workspace up to trial $b$ is a relational database defined as 
\[
\mathcal W^{(\leq b)}
=
\left\{
\widetilde{\mathcal D},
\mathcal T_{\mathrm{trials}}^{(\leq b)}, 
\mathcal T_{\mathrm{preds}}^{(\leq b)}
\right\}
\]
where $\widetilde{\mathcal D}$ is the augmented context database and for each trial  $j\in \{1,\dots,b\}$ the table \(\mathcal T_{\mathrm{trials}}^{(\leq b)}\) contains one trial-level record and \(\mathcal T_{\mathrm{preds}}^{(\leq b)}\) contains one row per evaluated target example with its validation score and ground truth value for each trial. 
After at least one validation, the agent can inspect the accumulated workspace via \texttt{query\_eval\_workspace()}, which executes evaluation queries against \(\mathcal{W}^{(\leq b_{\text{curr}})}\), where \(b_{\text{curr}}\) is the latest validation index. Prediction records are keyed by trial index and stable \texttt{row\_id}, with diagnostic metadata stored during validation. This lets the agent align outcomes across trials, inspect errors, aggregate diagnostics, and revise later feature programs, model configurations, or target transformations. See \Cref{app: evaluation_workspace_details} for further details.

\subsection{Inference phase}
\vspace{-0.1em}
After the search, the LLM agent is removed from the prediction pipeline.
Let $B$ be the total number of validate actions for the policy rollout. We then simply select the feature program and predictive model that achieved the highest validation score, i.e.,
\[
b^\star
:=
\argmax_{1 \leq b \leq B}\,
\mathrm{score}_{\mathrm{val}}(\vq^{(b)},\varphi^{(b)}),
\qquad
(\vq^\star,\varphi^\star)
:=
(\vq^{(b^\star)},\varphi^{(b^\star)}).
\]
In practice, we execute $K=5$ independent policy rollouts and select the candidate that achieves the highest validation score across all searches, which we refer to as \emph{cross-rollout model selection}. For test evaluation, the feature construction map $\Phi_{\vq^\star}$ induced by $\vq^\star$ is applied to train, validation and test tables, yielding $\mX_{\mathrm{train}}$, $\mX_{\mathrm{val}}$ and $\mX_{\mathrm{test}}$ and 
a predictor $f^\star$ -- initialized w.r.t the model configuration \(\varphi^\star\) -- is fitted to training data \((\mX_{\mathrm{train}},\vy_{\mathrm{train}})\) and validation data $(\mX_{\mathrm{val}},\vy_{\mathrm{val}})$. 
Finally, test predictions are computed as
\(
\hat{\vy}_{\mathrm{test}}
=
f^\star\!\left(
\mX_{\mathrm{test}}
\right).
\)

\subsection{Properties}
\label{sec:properties}
\vspace{-0.1em}

\noindent\textbf{Intrinsic interpretability.}
Each feature is defined by an explicit SQL query, making predictions directly inspectable through the selected queries and the downstream model's feature usage. We illustrate this in \Cref{sec:case_study}, where the agent discovers semantically meaningful feature blocks such as recency, post activity, and commenting behavior.

\noindent\textbf{Expressivity.}
Since \textsc{RelAgent} discovers SQL programs during search, its expressivity is bounded by the available query language $\mathcal{L}$ rather than a fixed architecture. SQL can express motifs such as triangle participation via self-joins and distinctness constraints, which standard Message Passing Neural Networks cannot. In \Cref{app:synthetic_triangle}, the agent discovers a filtered three-way self-join for directed triangle detection, illustrating how query-based predictors can represent logical relational patterns beyond fixed message-passing architectures, though not always within a fixed search budget. 

In contrast to automated feature generation approaches such as deep feature synthesis~\citep{kanter2015deep}, SQL can capture joint conditions across multiple rows of a child table rather than only column-wise aggregations, enabling strictly more expressive feature programs (see \Cref{app:dfs}).

\noindent\textbf{Query-induced invariance.}
Similar to expressivity, \textit{relational invariance}, i.e., invariance to permutations of row identifiers and table orderings, is induced by the query language $\mathcal{L}$, not a model architecture: as long as each selected SQL feature is relationally invariant, so is the resulting predictor. In our experiments queries that access row identifiers were not explicitly forbidden, however we found that in all final selected programs, only relationally invariant queries were chosen (see \Cref{app:exp_details}).

\section{Experiments}\label{sec:experiments}
\vspace{-0.1em}

We evaluate \textsc{RelAgent} along five axes: performance on classification and regression tasks, component ablations, backbone robustness, search-budget scaling (\Cref{app:search_budget}), and qualitative search behavior study (\Cref{sec:case_study}). We also include a discussion of Deep Feature Synthesis in \Cref{app:dfs}.

\subsection{Setup}
\vspace{-0.1em}

\noindent\textbf{Databases and tasks.}
We evaluate our method on three relational benchmarks, RelBenchV1~\citep{relbench}, RelBenchV2\footnote{We did not include MIMIC dataset due to licensing constraints.}~\citep{relbenchv2}, and 4DBInfer~\citep{wang20244dbinfer}. These benchmarks span a diverse set of domains, including e-commerce, social and Q\&A platforms, medicine, academia, and sports, and cover a range of temporal prediction tasks such as churn prediction, engagement prediction, and lifetime value (LTV) estimation. Full dataset statistics are provided in \Cref{tab:dataset_stats}.

\noindent\textbf{Baselines.}
We compare against four groups of baselines
: supervised tabular models, relational deep learning methods, relational foundation models, and LLM as feature engineers approaches:

\begin{itemize}[leftmargin=*, itemsep=2pt, topsep=2pt]
    \item \textbf{Supervised tabular models.}
    We include native tabular predictors and relational-to-tabular pipelines: XGBoost~\citep{chen2016xgboost}, LightGBM~\citep{lightgbm}, AutoGluon over flattened tables~\citep{autogluon}, and human feature-engineering pipelines such as DS+LightGBM, DS+AutoGluon, and DS+TabPFN-2.5~\citep{relbench,hudovernik2026kumorfm}.

    \item \textbf{Relational deep learning.}
    We include graph neural networks such as GraphSAGE \citep{graphsage}, GAT \citep{gat}, PNA \citep{pna}, and RelGNN \citep{relgnn}; graph transformers such as HGT/HGT$_{\text{PE}}$ \citep{hgt} and RelGT \citep{relgt}; and language-model-based relational predictors, including $\mathrm{LLM}_1$ \citep{ranjan2026rt}, $\mathrm{LLM}_2$ \citep{wydmuch24}.

    \item \textbf{Relational foundation models.}
    We compare against Griffin~\citep{griffin}, RT$_{\text{zero}}$~\citep{ranjan2026rt}, Rel-Zero~\citep{relllm}, KumoRFM-v1~\citep{fey2025kumorfm-v1}, and KumoRFM-v2~\citep{hudovernik2026kumorfm}. We also include hybrid or tabular foundation-model pipelines, including GNN+TabPFN-2.5~\citep{gnntabpfn}, RDBLearn~\citep{xu2026rdblearn}, and TabPFN-2.5~\citep{tabpfnv2}.
    
    \item \textbf{LLMs-as-feature-engineers.} We include FeatLLM~\citep{featllm} and ReFuGe~\citep{kim2026refuge} as LLM-based feature-synthesis baselines on RelBenchV1 and report FeatLLM results from~\citet{kim2026refuge}.
\end{itemize}

\begin{table}[t]
 \vspace{-2em}
\centering
\caption{\textbf{Test results on the entity classification tasks in RelBenchV1.} Higher is better (AUROC). We \textbf{bold} the best result and \underline{underline} the second best result.}
\label{tab:rbv1-cls}
\setlength{\tabcolsep}{1.3pt}
\footnotesize
\begin{tabular}{p{0.2cm}lR{0.75cm}R{0.75cm}R{0.75cm}R{0.75cm}R{0.75cm}R{0.75cm}R{0.75cm}R{0.75cm}R{0.75cm}R{0.75cm}R{0.75cm}R{0.75cm}|R{0.75cm}R{0.75cm}}
\toprule
&\multirow{2}{*}{\textbf{Method}} & \multicolumn{2}{c}{\texttt{\footnotesize f1}} & \multicolumn{2}{c}{\texttt{\footnotesize avito}} & \multicolumn{2}{c}{\texttt{\footnotesize event}} & \multicolumn{1}{c}{\texttt{\footnotesize trial}} & \multicolumn{2}{c}{\texttt{\footnotesize amazon}} & \multicolumn{2}{c}{\texttt{\footnotesize stack}} & \multicolumn{1}{c}{\texttt{\footnotesize hm}} & \multicolumn{1}{c}{\textbf{\footnotesize Avg}} & \multicolumn{1}{c}{\textbf{\footnotesize Rank}} \\
& &\multicolumn{1}{c}{\texttt{\tiny dnf}} & \multicolumn{1}{c}{\texttt{\tiny top3}} & \multicolumn{1}{c}{\texttt{\tiny click}} & \multicolumn{1}{c}{\texttt{\tiny visit}} & \multicolumn{1}{c}{\texttt{\tiny repeat}} & \multicolumn{1}{c}{\texttt{\tiny ignore}} & \multicolumn{1}{c}{\texttt{\tiny out}} & \multicolumn{1}{c}{\texttt{\tiny user}} & \multicolumn{1}{c}{\texttt{\tiny item}} & \multicolumn{1}{c}{\texttt{\tiny eng}} & \multicolumn{1}{c}{\texttt{\tiny badge}} & \multicolumn{1}{c}{\texttt{\tiny churn}} & \multicolumn{1}{c}{\scriptsize $\uparrow$} & \multicolumn{1}{c}{\scriptsize $\downarrow$} \\
\midrule
\multirow{4}{*}{\rotatebox{90}{\tiny Sup. Tabular}} & {\scriptsize LightGBM} & 68.56 & 73.92 & 53.60 & 53.05 & 68.04 & 79.93 & 70.09 & 52.22 & 62.54 & 63.39 & 63.43 & 55.21 & 63.66 & 18.25\\
& {\scriptsize DS+LightGBM} & 69.80 & 82.40 & 64.27 & 64.46 & 75.42 & 84.23 & \underline{72.00} & 67.60 & 81.80 & 90.30 & 86.20 & 69.00 & 75.62 & 8.42\\
&{\scriptsize DS+AutoGluon}& 70.01 & 82.53 & 65.66 & 65.87 & 74.85 & 82.31 & 70.51 & 68.12 & 80.45 & 89.94 & 85.38 & 68.71 & 75.36 & 9.25 \\
& {\scriptsize DS+TabPFN-2.5}& 71.16 & 77.70 & 64.21 & 64.53 & 76.43 & 85.70 & 70.75 & 66.28 & 79.83 & 89.07 & 85.08 & 68.31 & 74.92 & 10.25 \\
\midrule
\multirow{5}{*}{\rotatebox{90}{\tiny Sup. Relational}}& {\scriptsize GraphSAGE} & 72.62 & 75.54 & 65.90 & 66.20 &
76.89 & 81.62 & 68.60 & {70.42} & \textbf{82.81} & \underline{90.59} & \underline{88.86} & 69.88 & 75.83 & 6.75\\
& {\scriptsize HGT} & 70.77 & 70.75 & 63.76 & 64.32 & 64.96 & 82.47 & 58.37 & 66.43 & 77.97 & 88.47 & 86.08 & 66.95 & 71.78 & 13.58 \\
& {\scriptsize HGT$_{\textrm{PE}}$} & 71.17 & 76.27 & 64.57 & 64.95 & 65.36 & 81.61 & 59.21 & 66.19 & 78.03 & 88.17 & 85.66 & 65.69 & 72.24 & 12.75 \\
& {\scriptsize RelGNN} & 75.29 & 85.69 & 68.23 & 66.18 & 79.61 & 86.18 & 71.24 & \textbf{70.99} & {82.64} & \textbf{90.75} & \textbf{88.98} & \underline{70.93} & {78.06} & {3.83} \\
& {\scriptsize RelGT} & 75.87 & 83.52 & {68.30} & 66.78 & 76.09 & 81.57 & 68.61 & 70.39 & 82.55 & 90.53 & 86.32 & 69.27 & 76.65 & 6.38 \\
\midrule
\multirow{10}{*}{\rotatebox{90}{\tiny  Zero-Shot Foundational}} & {\scriptsize LLM$_1$} & 75.80 & \underline{91.40} & 59.80 & 62.70 & 71.40 & 69.30 & 57.40 & 58.10 & 62.10 & 78.00 & 80.00 & 59.80 & 68.82 & 15.04 \\
& {\scriptsize LLM$_2$} & 80.03 & 87.11 & 53.36 & 54.07 & 70.11 & 68.65 & 59.17 & 62.55 & 73.41 & 81.23 & 79.99 & 63.81 & 69.46 & 14.75 \\
& {\scriptsize Rel-Zero} & 71.84 & 70.64 & 62.28 & 56.17 & 68.12 & 61.32 & 59.02 & 60.07 & 64.10 & 69.46 & 62.12  & 55.95 & 63.42 & 17.67 \\
& {\scriptsize TabPFN-2.5}& 61.21 & 77.35 & 54.44 & 51.45 & 69.10 & 70.91 & 71.74 & 57.14 & 59.30 & 59.66 & 52.26 & 54.99 & 61.63 & 18.08 \\
& {\scriptsize Griffin}& 57.70 & 82.50 & 45.90 & 60.70 & 71.88 & 83.27 & 51.00 & 62.30 & 69.00 & 77.50 & 73.50 & 60.20 & 66.29 & 16.58 \\
& {\scriptsize RT$_{\textrm{zero}}$} & 81.20 & 89.30 & 59.50 & 61.80 & 73.22 & 77.47 & 51.80 &  64.00 & 70.90 & 75.70 & 80.10 & 62.80 & 70.65 & 14.08 \\
& {\scriptsize GNN+TabPFN} & 64.26 & 74.67 & 60.68 & 65.35 & 73.58 & 84.13 & 56.68 & 64.30 & 77.89 & 87.48 & 85.34 & 64.34 & 71.56 & 13.75 \\
& {\scriptsize RDBLearn} & 70.87 & 79.69 & \textbf{69.04} & 65.49 & 75.04 & 82.52 & 71.58 & 67.57 & 82.07 & 89.39 & 85.26 & 68.05 & 75.55 & 8.92 \\
& {\scriptsize KumoRFM-v1} & \underline{82.41} & 91.07 & 64.85 & 64.11 & 76.08 & \underline{89.20} & 70.79 & 67.29 & 79.93 & 87.09 & 80.00 & 67.71 & 76.71 & 8.71 \\
& {\scriptsize KumoRFM-v2} & \textbf{84.59} & \textbf{92.18} & 67.42 & \textbf{69.41} & \textbf{81.66} & \textbf{90.83} & \underline{72.03} & 69.10 & 82.17 & 89.40 & 87.15 & 69.27 & \textbf{79.60} & \underline{3.38} \\
\midrule
\multirow{3}{*}{\rotatebox{90}{\tiny LLM as FE.}}
& FeatLLM & 70.43 & 52.72 & 48.69 & 51.15 & 46.67 & 47.88 & 57.35 & - & 55.11 &  42.56 & - & - & - & - \\
& ReFuGe & 76.26 & 83.04 & 63.20 & 66.43 & 77.14  & 83.40 & \textbf{72.22} & - & 67.00 & 87.66 & 83.80 & 68.10 & - & - \\
& \scriptsize \textsc{RelAgent} & 78.34 & 85.23 & \underline{68.36} & \underline{67.79} & {78.20} & 87.25 & {71.86} & \underline{70.78} & \underline{82.84} & 90.41 & 88.42  & \textbf{71.07} & \underline{78.38} & \textbf{3.17} \\
\bottomrule
\end{tabular}
\end{table}

\noindent\textbf{Implementations \& evaluations.}
\textsc{RelAgent} uses GPT-5.2~\citep{openai2025gpt52} as the LLM backbone, CAMEL~\citep{camel} as the agent framework, and \textsc{DuckDB}~\citep{duckdb} as the SQL engine. The agent selects among seven tree-based predictors: LightGBM variants with standard boosting, DART, and GOSS~\citep{lightgbm}; Random Forest~\citep{breiman2001random}; XGBoost and XGBoost+DART~\citep{chen2016xgboost,gilad2015dart}; and CatBoost~\citep{prokhorenkova2017catboost}. We run five independent search rollouts and select the final model with the highest validation score. We report average AUROC for classification, normalized MAE for regression, and average rank over methods with complete results. See efficiency analysis in \Cref{app:efficiency} and the full experimental details in \Cref{app:exp_details}. Code is available at \url{https://github.com/HxyScotthuang/RelAgent}.

\subsection{Entity classification}
\vspace{-0.1em}

We evaluate \textsc{RelAgent} on entity classification tasks from RelBenchV1, 4DBInfer, and RelBenchV2, comparing against supervised tabular pipelines, supervised relational models, zero-shot relational foundation models, and LLM-as-feature-engineer. Results are shown in \Cref{tab:rbv1-cls,tab:dbinfer,tab:relbenchv2-binary}.

Observe that on RelBenchV1, \textsc{RelAgent} achieves the second-best average AUROC, reaching \(78.38\), behind only the proprietary relational foundation model KumoRFM-v2 (\(79.60\)). On the other hand, \textsc{RelAgent} obtains the best overall rank (\(3.17\)), indicating that its performance is consistently strong across tasks rather than isolated wins. 

\begin{wraptable}{r}{7.7cm}
\centering
\captionof{table}{\textbf{Test results on the entity classification tasks in RelBenchV2.} Higher is better (AUROC). We \textbf{bold} the best and \underline{underline} the second best result.}
\label{tab:relbenchv2-binary}
\setlength{\tabcolsep}{1pt}
\footnotesize
\begin{tabular}{lcccc|cc}
\toprule
 \multirow{2}{*}{\textbf{Method}}& \multicolumn{3}{c}{\texttt{ratebeer}} & \multicolumn{1}{c}{\texttt{arxiv}} & \textbf{Avg} & \textbf{Rank} \\
& \multicolumn{1}{c}{\texttt{beer}} & \multicolumn{1}{c}{\texttt{user}} & \multicolumn{1}{c}{\texttt{dormant}} & \multicolumn{1}{c}{\texttt{citation}} &  $\uparrow$ &  $\downarrow$ \\
\midrule
LightGBM & 76.21 & 83.92 & 75.79 & 71.21 & 76.78 & 4.75 \\
GraphSAGE & 78.67 & 94.27 & 80.51 & \underline{82.50} & 83.99 & 2.75 \\
KumoRFM-v1 & 75.06 & 91.38 & 77.10 & 80.62 & 81.04 & 4.25 \\
KumoRFM-v2 & \underline{83.84} & \underline{97.43} & \underline{80.65} & 81.71 & \underline{85.91} & \underline{2.25} \\
\midrule
\textsc{RelAgent} & \textbf{84.70} & \textbf{98.63} & \textbf{83.33} & \textbf{82.62} & \textbf{87.32} & \textbf{1.00} \\
\bottomrule
\end{tabular}
%
\centering
\captionof{table}{\textbf{Test results on the entity classification tasks in 4DBInfer.} Higher is better (AUROC). We \textbf{bold} the best and \underline{underline} the second best result. }
\label{tab:dbinfer}
\setlength{\tabcolsep}{1.5pt}
\footnotesize
\begin{tabular}{llrrrrr|rr}
\toprule
& \multirow{2}{*}{\textbf{Method}} & \multicolumn{1}{c}{\footnotesize AB} & \multicolumn{1}{c}{\footnotesize OB} & \multicolumn{1}{c}{\footnotesize RR} & \multicolumn{2}{c}{\footnotesize SE} & \multicolumn{1}{c}{\textbf{\footnotesize Avg}} & \multicolumn{1}{c}{\textbf{\footnotesize Rank}} \\
& & \multicolumn{1}{c}{\texttt{\scriptsize churn}} & \multicolumn{1}{c}{\texttt{\scriptsize ctr}} & \multicolumn{1}{c}{\texttt{\scriptsize cvr}} & \multicolumn{1}{c}{\texttt{\scriptsize vote}} & \multicolumn{1}{c}{\texttt{\scriptsize churn}} & \multicolumn{1}{c}{\scriptsize $\uparrow$} & \multicolumn{1}{c}{\scriptsize $\downarrow$} \\
\midrule
\multirow{4}{*}{\rotatebox{90}{\tiny  Sup. Tab.}}& {XGBoost} & 50.00 & 50.00 & 50.00 & 49.68 & 50.84 & 50.10 & 11.50 \\
& {AutoGluon} & 50.00 & 49.69 & 50.96 & 50.81 & 50.00 & 50.29 & 11.50 \\
& {DFS+XGBoost} & 69.22 & 54.21 & 79.06 & 86.75 & 82.51 & 74.35 & 10.00 \\
& {DFS+AutoGluon} & 72.91 & 54.94 & 80.08 & 88.49 & 83.96 & 76.08 & 8.40 \\
\midrule
\multirow{4}{*}{\rotatebox{90}{\tiny  Sup. Rel.}}& {GAT} & 76.22 & 61.46 & 82.84 & 88.53 & 86.45 & 79.10 & 6.40 \\
& {HGT} &77.30 & \underline{62.60} & \underline{84.95} & 88.17 & 86.70 & 79.94 & \underline{4.20} \\
& {PNA} & 76.45 & {62.49} & 83.67 & \underline{88.96} & 86.64 & 79.64 & 4.60 \\
& {GraphSAGE} & 75.71 & 62.39 & {84.70} & 88.61 & 85.58 & 79.40 & 5.20 \\
\midrule
\multirow{3}{*}{\rotatebox{90}{\tiny  Found.}}& {RDBLearn} & \underline{77.41} & 54.47 & 84.69 & 88.45 &\underline{87.96} & 78.60 & 5.00 \\
& {KumoRFM-v1} & 75.59 & 61.11 & 84.36 &88.92 & 87.87 & 79.17 & 5.50 \\
& {KumoRFM-v2} & 77.06 & 61.97 & 84.36 & 88.50 &87.92 & \underline{79.96} & 4.70 \\
\midrule
& \textsc{RelAgent} & \textbf{79.44} & \textbf{62.63} & \textbf{86.35} & \textbf{89.45} & \textbf{89.03} & \textbf{81.38} & \textbf{1.00} \\
\bottomrule
\end{tabular}
\vspace{-2em}
\end{wraptable}
\noindent\textbf{Supervised models.}
\textsc{RelAgent} substantially improves over LightGBM on raw tables and outperforms human feature-engineering pipelines with LightGBM, AutoGluon, and TabPFN-2.5. This suggests that the agent can automate relational feature engineering while remaining competitive with human-designed pipelines. \textsc{RelAgent} achieves the highest average AUROC and the best average rank, outperforming all relational supervised baselines. Thus, agentic search via SQL is competitive with end-to-end supervised relational architectures on this benchmark.

\noindent\textbf{Foundation models.}
\textsc{RelAgent} also outperforms all open-source zero-shot foundation-models in average AUROC. Only the closed-source KumoRFM-v2 achieves a higher average AUROC (\(79.60\) vs. \(78.38\)), while \textsc{RelAgent} obtains the best average rank (\(3.17\) vs. \(3.38\)). Unlike foundation models used directly as predictors, \textsc{RelAgent} uses the LLM only during search; inference is performed without further LLM calls.

\noindent\textbf{LLM as feature engineers.}
We also report FeatLLM and ReFuGe as LLM-based feature-engineering references. On overlapping tasks, \textsc{RelAgent} is competitive with or stronger than ReFuGe, while supporting evaluation-workspace queries and joint feature/model selection.

\noindent\textbf{\textsc{RelAgent} on RelBenchV2 and 4DBInfer.}
\textsc{RelAgent} achieves the best AUROC on all RelBenchV2 and 4DBInfer entity classification tasks, improving the averages from KumoRFM-v2 from \(85.91\) to \(87.32\) and from \(79.96\) to \(81.38\), respectively; see \Cref{tab:relbenchv2-binary,tab:dbinfer}. On 4DBInfer, it also outperforms DFS-based baselines, suggesting that guided SQL search can improve over generic relational aggregation. Unlike fixed DFS aggregations, SQL feature programs can express task-specific row-level interactions, such as co-occurrence predicates within a related table; see \Cref{app:dfs}.

\subsection{Entity regression}
\vspace{-0.1em}

\begin{table}[t]
\centering
\caption{\textbf{Test results on entity regression tasks in RelBenchV1.} Lower is better (MAE). Average
performance \bm{$\mu$}$_n$ is normalized relative to the LightGBM baseline. We \textbf{bold} the best result and \underline{underline} the second best result.}
\label{tab:rbv1-reg}
\footnotesize
\setlength{\tabcolsep}{3pt}
\begin{tabular}{p{0.2cm}lR{0.75cm}R{0.75cm}R{0.75cm}R{0.75cm}R{0.75cm}R{0.75cm}R{0.75cm}R{0.75cm}R{0.75cm}|R{0.75cm}R{0.75cm}}
\toprule
& \multirow{2}{*}{\textbf{Method}} & \multicolumn{1}{c}{\texttt{\footnotesize f1}} & \multicolumn{1}{c}{\texttt{\footnotesize avito}} & \multicolumn{1}{c}{\texttt{\footnotesize event}} & \multicolumn{2}{c}{\texttt{\footnotesize trial}} & \multicolumn{2}{c}{\texttt{\footnotesize amazon}} & \multicolumn{1}{c}{\texttt{\footnotesize stack}} & \multicolumn{1}{c}{\texttt{\footnotesize hm}} & \multicolumn{1}{c}{\bm{$\mu$}$_n$} & \multicolumn{1}{c}{\textbf{\footnotesize Rank}} \\
& & \multicolumn{1}{c}{\texttt{\tiny pos}} & \multicolumn{1}{c}{\texttt{\tiny ctr}} & \multicolumn{1}{c}{\texttt{\tiny attend}} & \multicolumn{1}{c}{\texttt{\tiny adverse}} & \multicolumn{1}{c}{\texttt{\tiny success}} & \multicolumn{1}{c}{\texttt{\tiny user}} & \multicolumn{1}{c}{\texttt{\tiny item}} & \multicolumn{1}{c}{\texttt{\tiny votes}} & \multicolumn{1}{c}{\texttt{\tiny sales}} & \multicolumn{1}{c}{\scriptsize $\downarrow$} & \multicolumn{1}{c}{\scriptsize $\downarrow$} \\
\midrule
& {\scriptsize Global Median} & 4.399 & 0.043 & 0.264 & 57.533 & 0.462 & 16.783 & 64.234 & 0.068 & 0.076 & 1.062 & 13.22 \\
& {\scriptsize Entity Median} & 8.519 & 0.046 & 0.269 & 57.930 & 0.441 & 17.423 & 66.436 & 0.069 & 0.078 & 1.190 & 15.06 \\
\midrule
\multirow{4}{*}{\rotatebox{90}{\tiny Sup. Tabular}}& {\scriptsize LightGBM} & 4.170 & 0.041 & 0.264 & 44.011 & 0.425 & 16.783 & 60.569 & 0.068 & 0.076 & 1.000 & 10.61 \\
& {\scriptsize DS+LightGBM} & 3.963 & 0.044 & 0.284 & \underline{40.581} & 0.407 & \underline{13.928} & \textbf{41.122} & \underline{0.065} & 0.036 & 0.879 & 5.56 \\
& {\scriptsize DS+AutoGluon} & 4.251 & 0.045 & 0.256 & 44.706 & 0.430 & 14.399 & 45.390 & 0.068 & 0.043 & 0.918 & 8.39 \\
& {\scriptsize DS+TabPFN-2.5} & 4.373 & 0.044 & 0.318 & 47.168 & 0.432 & 15.631 & 47.908 & 0.080 & 0.059 & 1.009 & 11.11 \\
\midrule
\multirow{5}{*}{\rotatebox{90}{\tiny Sup. Relational}}& {\scriptsize GraphSAGE} & 4.022 & 0.041 & 0.258 & 44.473 & 0.400 & 14.313 & 50.053 & \underline{0.065} & 0.056 & 0.918 & 7.17 \\
& {\scriptsize HGT} & 4.226 & 0.046 & 0.264 & 45.169 & 0.443 & 15.412 & 55.868 & 0.068 & 0.064 & 0.988 & 11.61 \\
& {\scriptsize HGT$_{\textrm{PE}}$} & 4.392 & 0.048 & 0.261 & 42.648 & 0.440 & 15.864 & 55.849 & 0.068 & 0.064 & 0.992 & 10.83 \\
& {\scriptsize RelGNN} & 3.798 & 0.037 & 0.238 & 44.461 & \textbf{0.301} & 14.230 & 48.767 & \underline{0.065} &  0.054 & 0.861 & 4.94 \\
& {\scriptsize RelGT} & 3.917 & 0.035 & 0.250 & 43.992 & \underline{0.326} & 14.267 & 48.922 & \underline{0.065} & 0.054 & 0.869 & 5.33 \\
\midrule
\multirow{6}{*}{\rotatebox{90}{\tiny  Zero-shot Foundational}}
& {\scriptsize TabPFN-2.5}& 4.446 & 0.044 & 0.527 & 55.187 & 0.469 & 17.513 & 60.996 & 0.104 & 0.096 & 1.258 & 15.78 \\
& {\scriptsize Griffin} & 4.460 & 0.050 & 0.461 & 78.232 & 0.463 & 35.590 & 53.214 & 0.092 & 0.151 & 1.471 & 16.44 \\
& {\scriptsize RT$_{\textrm{zero}}$} & 2.901 & 0.058 & 0.379 & 73.999 & 0.455 & 18.802 & 57.996 & 0.110 & 0.089 & 1.240 & 14.89 \\
& {\scriptsize RDBLearn} & 3.834 & \underline{0.034} & \underline{0.237} & 43.913 & 0.424 & 14.540 & 48.559 & 0.068 & 0.064 & 0.906 & 6.33 \\
& {\scriptsize KumoRFM-v1} & \textbf{2.747} & 0.035 & 0.264 & 58.231 & 0.417 & 16.161 & 55.254 & \underline{0.065} & 0.040 & 0.908 & 7.78 \\
& {\scriptsize KumoRFM-v2} & \underline{2.854} & \underline{0.034} & \textbf{0.235} & 43.293 & 0.433 & \textbf{13.921} & 46.992 & \textbf{0.064} & \textbf{0.034} & \underline{0.822} & \underline{3.11} \\
\midrule
& \scriptsize\textsc{RelAgent} & 4.019 & \textbf{0.033} & 0.241 & \textbf{37.194} & 0.386 & 13.949 & \underline{41.765} & \textbf{0.064} & \underline{0.035} & \textbf{0.817} & \textbf{2.83} \\
\bottomrule
\end{tabular}
\end{table}

\begin{wraptable}{r}{7cm}
\centering
\vspace{-1em}
\caption{\textbf{Test results on the regression tasks in RelBenchV2.} Lower is better (MAE). Average performance $\mu_n$ is normalized relative to the LightGBM baseline. We \textbf{bold} the best and \underline{underline} the second best result among all models.}

\label{tab:relbenchv2-reg}
\setlength{\tabcolsep}{1.5pt}
\footnotesize
\begin{tabular}{lcc|cc}
\toprule
\multirow{2}{*}{\textbf{Method}} & \multicolumn{1}{c}{\texttt{ratebeer}} & \multicolumn{1}{c}{\texttt{arxiv}} & \textbf{$\mu_n$} & \textbf{Rank}\\
& \multicolumn{1}{c}{\texttt{user-count}} & \multicolumn{1}{c}{\texttt{publication}} & \scriptsize $\downarrow$ & $\downarrow$ \\
\midrule
Global Median & 15.124 & 0.577 & 0.872 & 5.50 \\
Entity Median & 13.079 & 0.874 & 1.079 & 5.50 \\
LightGBM & 20.350 & 0.577 & 1.000 & 6.00 \\
GraphSAGE & 7.374 & 0.513 & 0.626 & 3.00 \\
KumoRFM-v1 & 11.063 & 0.518 & 0.721 & 4.00 \\
KumoRFM-v2 & \underline{7.298} & \underline{0.487} & \underline{0.601} & \underline{2.00} \\
\midrule
\textsc{RelAgent} & \textbf{6.021} & \textbf{0.462} & \textbf{0.548} & \textbf{1.00} \\
\bottomrule
\end{tabular}
\vspace{-1em}
\end{wraptable}

We evaluate \textsc{RelAgent} on entity regression tasks in RelBenchV1 and RelBenchV2; results are shown in \Cref{tab:rbv1-reg,tab:relbenchv2-reg}. \textsc{RelAgent} achieves the best normalized average performance and average rank on RelBenchV1, improving over KumoRFM-v2, and performs best on both RelBenchV2 regression tasks.

We attribute this to the flexibility of program-level search. Regression tasks in these benchmarks often require task-specific choices beyond relational aggregation: targets such as lifetime value and sales are often heavy-tailed, and different tasks benefit from different temporal windows, model families, and objectives. \textsc{RelAgent} searches over these choices jointly. In the selected programs, the agent consistently chooses \(L_1\) objectives and applies log-target transformations in seven of nine RelBenchV1 tasks, while also selecting task-specific SQL features and model configurations. This suggests that the gains come from optimizing the full predictive program, rather than from SQL feature construction alone.

\subsection{Ablations}
\label{sec:ablations}

We perform ablations to isolate the contribution of iterative feedback, semantic schema information, evaluation workspace, as well as the downstream supervised predictor and backbone LLM choice. Unless otherwise stated, all ablations use the same evaluation protocol as the main experiments and report the same task-level metrics aggregated across datasets. We report the averaged ablation results in \Cref{tab:ablation-rlv}, and the full results in \Cref{tab:ablation-rlv-full}.

\begin{wraptable}{r}{4.2cm}
\centering
\vspace{-1em}
\caption{\textbf{Ablation results on the entity classification tasks in RelBenchV1.} Higher is better (AUROC). }
\label{tab:ablation-rlv}
\setlength{\tabcolsep}{6pt}
\footnotesize
\begin{tabular}{@{}lr@{}}
\toprule
\textbf{Model Variants} & \textbf{Avg} $\uparrow$ \\
\midrule
\textsc{RelAgent} & \textbf{78.38} \\
\quad w/o feedback & 66.86 \\
\quad w/o eval workspace & 75.78 \\
\quad w/o semantic & 74.44 \\
\midrule
\quad rule-only & 69.92 \\
\quad catboost-only & 75.95 \\
\quad lightgbm-only & 75.94 \\
\quad xgboost-only & 73.42 \\
\quad xgboost-dart-only & 74.97 \\
\quad tabpfn-v2.5-only & 78.13 \\
\midrule
\quad with Deepseek-V4 & 76.00 \\
\quad with Qwen3-32B & 70.13 \\
\bottomrule
\end{tabular}
\vspace{-1em}
\end{wraptable}
\noindent\textbf{Iterative feedback loop.}
To isolate the value of iterative validation and revision, we compare \textsc{RelAgent} with a variant that does not allow program validation tools and is asked to output the final program and predictive model configuration directly. Removing this loop reduces average AUROC from \(78.38\) to \(66.86\), suggesting that the gains come not only from SQL features, but also from repeated validation, where the agent observes execution outcomes, debugs failures, and revises subsequent feature programs and model choices.

\noindent\textbf{Evaluation workspace.}
We compare \textsc{RelAgent} with a variant that receives validation feedback but cannot access the accumulated workspace \(\mathcal W^{(\leq b)}\). Removing this access reduces average AUROC from \(78.38\) to \(75.78\), showing that the workspace provides signal beyond scalar validation scores. In the \texttt{rel-stack/eng} case study in \Cref{sec:case_study}, the agent uses workspace diagnostics to identify constant or low-SHAP vote features and replaces them with voting-breadth features, such as the number of distinct posts voted on within recent windows. This illustrates how the workspace supports targeted error analysis.

\noindent\textbf{Semantic schema information.}
To assess the role of database semantics, we compare against an anonymized-schema variant where dataset names, table names, column names, and task descriptions are replaced by synthetic identifiers. This reduces average AUROC from \(78.38\) to \(74.44\). A closer look at the trace suggests that semantic clues help the agent identify task-relevant features such as recency, activity frequency, trends, and relational paths; without them, it falls back more often to generic counts, missingness statistics, and broad aggregates. The anonymized variant still performs reasonably well, indicating that tool-grounded search can recover useful relational signals even without human-readable schema semantics.

\noindent\textbf{Predictor choice.}
We compare \textsc{RelAgent} with fixed-predictor variants while keeping SQL feature search unchanged. Restricting the predictor to CatBoost, LightGBM, XGBoost, or XGBoost-DART reduces average AUROC from $78.38$ to $75.95$, $75.94$, $73.42$, and $74.97$, respectively. Notably, TabPFN-2.5 achieves $78.13$, approaching the performance of the full system with a fixed downstream predictor. A rule-only variant, where the agent writes Python rules to combine SQL-derived features without fitting a supervised predictor, achieves only $69.92$. These results indicate that task-specific SQL feature discovery should be paired with an effective learned downstream predictor.

\begin{wraptable}{r}{6cm}
\centering
\small
\caption{\textbf{SQL feature analysis.}
(a) Average AUROC on RelBenchV1 classification with
a fixed TabPFN-2.5 predictor.
(b) Test AUROC under restricted SQL operations on
three representative tasks.}
\label{tab:sql_feature_analysis}
\textbf{(a) SQL feature discovery}\\[3pt]
\begin{tabular}{lc}
\toprule
Feature representation & AUROC $\uparrow$ \\
\midrule
Entity-table only & 61.63 \\
Human-curated & 74.92 \\
DFS & 75.55 \\
\textsc{RelAgent} & \textbf{78.13} \\
\bottomrule
\end{tabular}

\vspace{0.8em}

\textbf{(b) SQL expressiveness ablations}\\[3pt]
\scriptsize
\setlength{\tabcolsep}{3pt}
\begin{tabular}{lcccc}
\toprule
Task & Full & 1-hop & No temporal & No cond. \\
\midrule
hm-churn & \textbf{71.07} & 70.47 & 68.34 & 69.10 \\
amazon-user & \textbf{70.78} & 69.33 & 64.69 & 70.43 \\
amazon-item & \textbf{82.84} & 82.52 & 73.90 & 82.08 \\
\bottomrule
\end{tabular}

\vspace{-1em}
\end{wraptable}

\noindent\textbf{Backbone robustness.} We evaluate alternative LLM backbones by rerunning the full pipeline with the same tools, search budget, and evaluation protocol. We choose DeepSeek-V4-Pro~\citep{deepseekv4} as a strong lower-cost backbone for token-intensive search. We also evaluate Qwen3-32B~\citep{qwen3} as a much smaller open-source backbone that can be hosted locally. DeepSeek-V4 reaches \(76.00\) average AUROC, close to the GPT-5.2 result of \(78.38\), while Qwen3-32B reaches \(70.13\).  Thus, stronger backbones improve search quality, while the tool-augmented design remains effective with smaller models.

\noindent\textbf{SQL feature discovery.}
We evaluate the quality of SQL features discovered by \textsc{RelAgent} while fixing the downstream predictor to TabPFN-2.5, shown in \Cref{tab:sql_feature_analysis} (a). Compared with Deep Feature Synthesis (DFS), human-curated features, and entity-table-only features, \textsc{RelAgent} achieves the highest average AUROC of $78.13$, outperforming DFS ($75.55$) and human-curated features ($74.92$) by $2.58$ and $3.21$ points, respectively. This demonstrates that the gains are not solely attributable to downstream predictor selection, but also arise from discovering task-specific relational features.

\noindent\textbf{SQL expressiveness.}
We investigate which relational operations contribute to \textsc{RelAgent}'s predictive performance by restricting the SQL search space to one-hop joins, removing temporal features, or disabling conditional aggregations. As shown in \Cref{tab:sql_feature_analysis} (b), removing temporal features causes the largest performance degradation across all three tasks, reducing AUROC by $2.73$, $6.09$, and $8.94$ points, respectively. In contrast, restricting join depth or conditional aggregations leads to smaller degradations. These results suggest that task-specific temporal feature construction is a particularly important benefit of SQL program synthesis.

\section{Conclusion}
\vspace{-0.1em}

We introduced \textsc{RelAgent}, an LLM-agent framework for relational learning that searches over SQL feature programs and downstream tabular predictors. By keeping relational structure external and executable, \textsc{RelAgent} uses the LLM only during search; the deployed predictor consists only of SQL feature construction and a standard predictive model. This yields fast, deterministic, and inspectable predictions without LLM calls at inference time.

Empirically, \textsc{RelAgent} achieves competitive or state-of-the-art results across classification and regression tasks, often matching or exceeding supervised relational models and relational foundation-model baselines. These results suggest that program-level search is a strong alternative to fixed relational architectures, especially when tasks require task-specific joins, temporal windows, objectives, or target transformations. Its main limitations are search cost and imperfect search reliability: the agent may miss useful deep joins, high-order relational patterns, or features from long free-text fields within a fixed budget; see \Cref{app:limitations}. Overall, \textsc{RelAgent} shows that LLMs can construct executable relational learning programs rather than serve only as direct predictors.

\section*{Acknowledgment}
This work was supported in part by the National Research Foundation of Korea (RS-2024-00351212) funded by the Korean government (MSIT). AITHYRA is supported by the Austrian Academy of Sciences and the not-for-profit Boehringer Ingelheim Stiftung.

\bibliographystyle{unsrtnat}  
\bibliography{refs}

@inproceedings{chen2016xgboost,
  title={Xgboost: A scalable tree boosting system},
  author={Chen, Tianqi and Guestrin, Carlos},
  booktitle={Proceedings of the 22nd acm sigkdd international conference on knowledge discovery and data mining},
  year={2016}
}

@inproceedings{lightgbm,
  title={{LightGBM}: A highly efficient gradient boosting decision tree},
  author={Ke, G. and Meng, Q. and Finley, T. and Wang, T. and Chen, W. and Ma, W. and Ye, Q. and Liu, T.},
  booktitle={NeurIPS},
  year={2017},
}

@article{autogluon,
  title={{AutoGluon-Tabular}: Robust and Accurate AutoML for Structured Data},
  author={Erickson, N. and Mueller, J. and Shirkov, A. and Zhang, H. and Larroy, P. and Li, M. and Smola, A.},
  journal={CoRR},
  year={2020},
}

@inproceedings{relbench,
  title={{RelBench}: A Benchmark for Deep Learning on Relational Databases},
  author={Robinson, J. and Ranjan, R. and Hu, W. and Huang, K. and Han, J. and Dobles, A. and Fey, M. and Lenssen, J. E. and Yuan, Y. and Zhang, Z. and He, X. and Leskovec, J.},
  booktitle={NeurIPS},
  year={2024},
}

@inproceedings{kanter2015deep,
  title={Deep feature synthesis: Towards automating data science endeavors},
  author={Kanter, James Max and Veeramachaneni, Kalyan},
  booktitle={2015 IEEE international conference on data science and advanced analytics (DSAA)},
  year={2015},
  organization={IEEE}
}

@inproceedings{graphsage,
  title={Inductive Representation Learning on Large Graphs},
  author={Hamilton, W. and Ying, Z. and Leskovec, J.},
  booktitle={NeurIPS},
  year={2017},
}

@inproceedings{gat,
  title={Graph Attention Networks},
  author={Veli{\v{c}}kovi{\'{c}}, P. and Cucurull, G. and Casanova, A. and Romero, A. and Li{\`{o}}, P. and Bengio, Y.},
  booktitle={ICLR},
  year={2018},
}

@inproceedings{pna,
  title={Principal Neighbourhood Aggregation for Graph Nets},
  author={Corso, G. and Cavalleri, L. and Beaini, D. and Li{\`{o}}, P. and Veli{\v{c}}kovi{\'{c}}, P.},
  booktitle={NeurIPS},
  year={2020},
}

@inproceedings{
    relgt,
    title={Relational Graph Transformer},
    author={Vijay Prakash Dwivedi and Sri Jaladi and Yangyi Shen and Federico Lopez and Charilaos I. Kanatsoulis and Rishi Puri and Matthias Fey and Jure Leskovec},
    booktitle={ICLR},
    year={2026},
}

@inproceedings{relgnn,
  title={Relgnn: Composite message passing for relational deep learning},
  author={Chen, Tianlang and Kanatsoulis, Charilaos and Leskovec, Jure},
  booktitle={ICML},
  year={2025}
}

@inproceedings{relllm,
  title={Large language models are good relational learners},
  author={Wu, Fang and Dwivedi, Vijay Prakash and Leskovec, Jure},
  booktitle={ACL},
  year={2025}
}

@inproceedings{
    ranjan2026rt,
    title={Relational Transformer: Toward Zero-Shot Foundation Models for Relational Data},
    author={Rishabh Ranjan and Valter Hudovernik and Mark Znidar and Charilaos I. Kanatsoulis and Roshan Reddy Upendra and Mahmoud Mohammadi and Joe Meyer and Tom Palczewski and Carlos Guestrin and Jure Leskovec},
    booktitle={ICLR},
    year={2026},
}

@article{tabpfnv2,
  title={Accurate predictions on small data with a tabular foundation model},
  author={Hollmann, N. and M{\"u}ller, S. and Purucker, L. and Krishnakumar, A. and K{\"o}rfer, M. and Hoo, S. B. and Schirrmeister, R. T. and Hutter, F.},
  journal={Nature},
  year={2025},
}

@inproceedings{
    griffin,
    title={Griffin: Towards a Graph-Centric Relational Database Foundation Model},
    author={Yanbo Wang and Xiyuan Wang and Quan Gan and Minjie Wang and Qibin Yang and David Wipf and Muhan Zhang},
    booktitle={ICML},
    year={2025},
}

@misc{fey2025kumorfm-v1,
      title={KumoRFM: A foundation model for in-context learning on relational data.}, 
      author={Fey, Matthias and Kocijan, Vid and Lopez, Federico and Lenssen, Jan Eric and Leskovec,Jure},
      booktitle={Online Tech Report},
      year={2025},
}

@article{wydmuch24,
  title={Tackling prediction tasks in relational databases with LLMs}, 
  author={Wydmuch, M. and Borchmann, L. and Graliński, F.},
  year={2024},
  volume={2411.11829},
  journal={CoRR},
}

@article{hudovernik2026kumorfm,
  title={KumoRFM-2: Scaling Foundation Models for Relational Learning},
  author={Hudovernik, Valter and L{\'o}pez, Federico and Kocijan, Vid and Nitta, Akihiro and Lenssen, Jan Eric and Leskovec, Jure and Fey, Matthias},
  journal={arXiv preprint arXiv:2604.12596},
  year={2026}
}

@inproceedings{hgt,
    author = {Hu, Ziniu and Dong, Yuxiao and Wang, Kuansan and Sun, Yizhou},
    title = {Heterogeneous Graph Transformer},
    year = {2020},
    booktitle = {Proceedings of The Web Conference 2020},
}

@inproceedings{gnntabpfn,
  title     = {Relational In-Context Learning on Structured Data via Neighborhood Aggregation and Structural Information},
  author    = {Meyer, Joe and Palczewski, Tom and Shaikh, Afreen and Mohammadi, Mahmoud and Ramprasath, Dinesh K. and Paresh, Karan and Reddy, Roshan U.  and Li, Mark},
  booktitle = {Proceedings of the AAAI-26 Summer Symposium Series: AI in Business: Intelligent Transformation and Management},
  year      = {2026},
}

@article{xu2026rdblearn,
  title={No Need to Train Your RDB Foundation Model},
  author={Xu, Linjie and Zhang, Yanlin and Gan, Quan and Wang, Minjie and Wipf, David},
  journal={arXiv preprint arXiv:2602.13697},
  year={2026}
}

@inproceedings{wang20244dbinfer,
  title={4DBInfer: A 4d benchmarking toolbox for graph-centric predictive modeling on RDBs},
  author={Wang, Minjie and Gan, Quan and Wipf, David and Cai, Zhenkun and Li, Ning and Tang, Jianheng and Zhang, Yanlin and Zhang, Zizhao and Mao, Zunyao and Song, Yakun and others},
  booktitle={NeurIPS},
  year={2024}
}

@misc{openai2025gpt52,
  author = {{OpenAI}},
  title = {Introducing GPT-5.2},
  year = {2025},
  howpublished = {https://openai.com/index/introducing-gpt-5-2/},
  note = {Accessed: 2026-04-22}
}

@inproceedings{camel,
  title={Camel: Communicative agents for" mind" exploration of large language model society},
  author={Li, Guohao and Hammoud, Hasan and Itani, Hani and Khizbullin, Dmitrii and Ghanem, Bernard},
  booktitle={NeurIPS},
  year={2023}
}

@Manual{duckdb,
  title = {duckdb: DBI Package for the DuckDB Database Management System},
  author = {Hannes Mühleisen and Mark Raasveldt},
  year = {2026},
}

@inproceedings{gilad2015dart,
  title={Dart: Dropouts meet multiple additive regression trees},
  author={Gilad-Bachrach, R and Rashmi, K},
  booktitle={AIStats},
  year={2015}
}

@article{breiman2001random,
  title={Random forests},
  author={Breiman, Leo},
  journal={Machine learning},
  year={2001},
  publisher={Springer}
}

@article{prokhorenkova2017catboost,
  title={CatBoost: Unbiased boosting with categorical features.},
  author={Prokhorenkova, Liudmila and Gusev, Gleb and Vorobev, Aleksandr and Dorogush, Anna Veronika and Gulin, Andrey},
  journal={arXiv preprint arXiv:1706.09516},
  year={2017}
}

@inproceedings{kim2026refuge,
  title={ReFuGe: Feature Generation for Prediction Tasks on Relational Databases with LLM Agents},
  author={Kim, Kyungho and Lee, Geon and Kim, Juyeon and Choi, Dongwon and Kang, Shinhwan and Shin, Kijung},
  booktitle={Proceedings of the ACM Web Conference},
  year={2026}
}

@inproceedings{tabert,
  title={TaBERT: Pretraining for joint understanding of textual and tabular data},
  author={Yin, Pengcheng and Neubig, Graham and Yih, Wen-tau and Riedel, Sebastian},
  booktitle={ACL},
  year={2020}
}

@inproceedings{react,
  title={ReAct: Synergizing Reasoning and Acting in Language Models},
  author={Yao, Shunyu and Zhao, Jeffrey and Yu, Dian and Shafran, Izhak and Narasimhan, Karthik R and Cao, Yuan},
  booktitle={NeurIPS 2022 Foundation Models for Decision Making Workshop},
  year={2022}
}

@inproceedings{tapex,
  title={TAPEX: Table Pre-training via Learning a Neural SQL Executor},
  author={Liu, Qian and Chen, Bei and Guo, Jiaqi and Ziyadi, Morteza and Lin, Zeqi and Chen, Weizhu and Lou, Jian-Guang},
  booktitle={ICLR},
  year={2022}
}

@inproceedings{tabllm,
  title={Tabllm: Few-shot classification of tabular data with large language models},
  author={Hegselmann, Stefan and Buendia, Alejandro and Lang, Hunter and Agrawal, Monica and Jiang, Xiaoyi and Sontag, David},
  booktitle={International Conference on Artificial Intelligence and Statistics},
  year={2023},
}

@inproceedings{dinsql,
  title={Din-sql: Decomposed in-context learning of text-to-sql with self-correction},
  author={Pourreza, Mohammadreza and Rafiei, Davood},
  booktitle={NeurIPS},
  year={2023}
}

@inproceedings{toolformer,
  title={Toolformer: Language models can teach themselves to use tools},
  author={Schick, Timo and Dwivedi-Yu, Jane and Dess{\`\i}, Roberto and Raileanu, Roberta and Lomeli, Maria and Hambro, Eric and Zettlemoyer, Luke and Cancedda, Nicola and Scialom, Thomas},
  booktitle={NeurIPS},
  year={2023}
}

@inproceedings{llaga,
  title={LLaGA: Large Language and Graph Assistant},
  author={Chen, Runjin and Zhao, Tong and Jaiswal, Ajay Kumar and Shah, Neil and Wang, Zhangyang},
  booktitle={ICML},
  year={2024},
}

@inproceedings{
reflexion,
title={Reflexion: language agents with verbal reinforcement learning},
author={Noah Shinn and Federico Cassano and Ashwin Gopinath and Karthik R Narasimhan and Shunyu Yao},
booktitle={NeurIPS},
year={2023},
}

@inproceedings{finkelshteinactions,
  title={Actions Speak Louder than Prompts: A Large-Scale Study of LLMs for Graph Inference},
  author={Finkelshtein, Ben and Cucerzan, Silviu and Jauhar, Sujay Kumar and White, Ryen W},
  booktitle={ICLR},
  year={2026},
}

@inproceedings{graphtoolinstruction,
  title={Graphtool-instruction: Revolutionizing graph reasoning in llms through decomposed subtask instruction},
  author={Wang, Rongzheng and Liang, Shuang and Chen, Qizhi and Zhang, Jiasheng and Qin, Ke},
  booktitle={KDD},
  year={2025}
}

@article{graphtoken,
  title={Let your graph do the talking: Encoding structured data for llms},
  author={Perozzi, Bryan and Fatemi, Bahare and Zelle, Dustin and Tsitsulin, Anton and Kazemi, Mehran and Al-Rfou, Rami and Halcrow, Jonathan},
  journal={arXiv preprint arXiv:2402.05862},
  year={2024}
}

@inproceedings{
talklikeagraph,
title={Talk like a Graph: Encoding Graphs for Large Language Models},
author={Bahare Fatemi and Jonathan Halcrow and Bryan Perozzi},
booktitle={ICLR},
year={2024},
}

@inproceedings{wang2023plan,
  title={Plan-and-solve prompting: Improving zero-shot chain-of-thought reasoning by large language models},
  author={Wang, Lei and Xu, Wanyu and Lan, Yihuai and Hu, Zhiqiang and Lan, Yunshi and Lee, Roy Ka-Wei and Lim, Ee-Peng},
  booktitle={ACL},
  year={2023}
}

@inproceedings{graphgpt,
  title={GraphGPT: graph instruction tuning for large language models},
  author={Tang, Jiabin and Yang, Yuhao and Wei, Wei and Shi, Lei and Su, Lixin and Cheng, Suqi and Yin, Dawei and Huang, Chao},
  year={2024},
  booktitle={SIGIR}
}

@article{instructglm,
  title={Language is All a Graph Needs},
  author={Ye, Ruosong and Zhang, Caiqi and Wang, Runhui and Xu, Shuyuan and Zhang, Yongfeng},
  journal={EACL},
  year={2024}
}

@inproceedings{graphicl,
  title={Graphicl: Unlocking graph learning potential in llms through structured prompt design},
  author={Sun, Yuanfu and Ma, Zhengnan and Fang, Yi and Ma, Jing and Tan, Qiaoyu},
  booktitle={Findings of the Association for Computational Linguistics: NAACL 2025},
  year={2025}
}

@inproceedings{caafe,
  title={Large language models for automated data science: Introducing caafe for context-aware automated feature engineering},
  author={Hollmann, Noah and M{\"u}ller, Samuel and Hutter, Frank},
  booktitle={NeurIPS},
  year={2023}
}

@inproceedings{featllm,
  title={Large Language Models Can Automatically Engineer Features for Few-Shot Tabular Learning},
  author={Han, Sungwon and Yoon, Jinsung and Arik, Sercan O and Pfister, Tomas},
  booktitle={ICML},
  year={2024},
}

@inproceedings{
octree,
title={Optimized Feature Generation for Tabular Data via {LLM}s with Decision Tree Reasoning},
author={Jaehyun Nam and Kyuyoung Kim and Seunghyuk Oh and Jihoon Tack and Jaehyung Kim and Jinwoo Shin},
booktitle={NeurIPS},
year={2024},
}

@article{abhyankar2025llmfe,
  title={Llm-fe: Automated feature engineering for tabular data with llms as evolutionary optimizers},
  author={Abhyankar, Nikhil and Shojaee, Parshin and Reddy, Chandan K},
  journal={arXiv preprint arXiv:2503.14434},
  year={2025}
}

@article{qwen3,
  title={Qwen3 technical report},
  author={Yang, An and Li, Anfeng and Yang, Baosong and Zhang, Beichen and Hui, Binyuan and Zheng, Bo and Yu, Bowen and Gao, Chang and Huang, Chengen and Lv, Chenxu and others},
  journal={arXiv preprint arXiv:2505.09388},
  year={2025}
}

@misc{deepseekv4,
  title        = {DeepSeek-V4: Towards Highly Efficient Million-Token Context Intelligence},
  author       = {{DeepSeek-AI}},
  year         = {2026},
  howpublished = {\url{https://huggingface.co/deepseek-ai/DeepSeek-V4-Pro/blob/main/DeepSeek_V4.pdf}},
  note         = {Technical report}
}

@article{relbenchv2,
  title={RelBench v2: A Large-Scale Benchmark and Repository for Relational Data},
  author={Gu, Justin and Ranjan, Rishabh and Kanatsoulis, Charilaos and Tang, Haiming and Jurkovic, Martin and Hudovernik, Valter and Znidar, Mark and Chaturvedi, Pranshu and Shroff, Parth and Li, Fengyu and others},
  journal={arXiv preprint arXiv:2602.12606},
  year={2026}
}

@article{codd1970relational,
  title={A relational model of data for large shared data banks},
  author={Codd, Edgar F},
  journal={Communications of the ACM},
  year={1970},
}

@misc{gpt_5.2_cost,
  author       = {{OpenAI}},
  title        = {{GPT-5.2} {API} Pricing},
  howpublished = {\url{https://openai.com/api/pricing/}},
  year         = {2026},
}

@misc{deepseek-v4-pro-cost,
  author       = {{DeepSeek}},
  title        = {{DeepSeek} {API} Pricing},
  howpublished = {\url{https://api-docs.deepseek.com/quick_start/pricing}},
  year         = {2026},
}

@inproceedings{ultra,
  title={Towards Foundation Models for Knowledge Graph Reasoning},
  author={Mikhail Galkin and Xinyu Yuan and Hesham Mostafa and Jian Tang and Zhaocheng Zhu},
  year={2024},
  booktitle = {ICLR},
}

@inproceedings{motif,
      title={How Expressive are Knowledge Graph Foundation Models?}, 
      author={Xingyue Huang and Pablo Barceló and Michael M. Bronstein and İsmail İlkan Ceylan and Mikhail Galkin and Juan L Reutter and Miguel Romero Orth},
      year={2025},
      booktitle={ICML},
}

@inproceedings{
trix,
title={{TRIX}: A More Expressive Model for Zero-shot Domain Transfer in Knowledge Graphs},
author={Yucheng Zhang and Beatrice Bevilacqua and Mikhail Galkin and Bruno Ribeiro},
booktitle={LoG},
year={2024},
}

@inproceedings{hyper,
  title={HYPER: A Foundation Model for Inductive Link Prediction with Knowledge Hypergraphs},
  author={Huang, Xingyue and Galkin, Mikhail and Bronstein, Michael M and Ceylan, Ismail Ilkan},
  booktitle={ICLR},
  year={2026},
}

@inproceedings{flock,
  title={Flock: A Knowledge Graph Foundation Model via Learning on Random Walks},
  author={Kim, Jinwoo and Huang, Xingyue and Olejniczak, Krzysztof and Min, Kyungbin and Bronstein, Michael and Hong, Seunghoon and {\.I}lkan Ceylan, {\.I}smail},
  booktitle={ICLR},
  year={2026},
}

\clearpage


\appendix

\section{Evaluation workspace details}
\label[appendix]{app: evaluation_workspace_details}

The \textit{evaluation workspace} is a persistent \textsc{DuckDB} database that accumulates trial results across the agent's search phase.
After each call to \texttt{validate\_program()}, the harness writes the outcome to the workspace; the agent reads it back through \texttt{query\_eval\_workspace(sql)}, which executes any read-only SQL query against the accumulated records.

\paragraph{Core tables.}
The base workspace maintains two tables, illustrated in \Cref{tab:eval_workspace_schema}\footnote{Table shows key fields; the full schema additionally includes \texttt{parent\_trial\_id}, \texttt{created\_at}, \texttt{resolved\_model\_config}, \texttt{notes} in \texttt{trials} and \texttt{eval\_cutoff} in \texttt{eval\_predictions}.}.
The \texttt{trials} table stores one record per validation trial, capturing the trial identifier, the names and a stable content hash of the submitted SQL feature blocks, the chosen downstream model, the prediction split, the primary validation metric, and a JSON serialization of all computed metrics.
The \texttt{eval\_predictions} table stores one row per evaluated target example, recording the entity identifier, ground-truth label, model output score, and optionally predicted class.
The key design choice is that \texttt{row\_id} is positionally stable across trials on the same split: it is the integer index of the row in the validation target table, assigned once and reused.
This allows the agent to join two trials directly on \texttt{row\_id} and compare predictions on the same examples.

A representative diagnostic query used by the agent that compares entity-level errors between two trials and finds examples of the 3rd trial fixed is:

\begin{lstlisting}[language=SQL, basicstyle=\ttfamily\footnotesize, frame=single]
SELECT p1.entity_id, p1.label, p1.score AS score_v1, p3.score AS score_v3
FROM eval_predictions p1
JOIN eval_predictions p3 ON p1.row_id = p3.row_id
WHERE p1.trial_id = 'val_0001' AND p3.trial_id = 'val_0003'
  AND ABS(p3.score - CAST(p3.label AS DOUBLE)) 
    < ABS(p1.score - CAST(p1.label AS DOUBLE))
ORDER BY ABS(p1.score - CAST(p1.label AS DOUBLE)) DESC LIMIT 20
\end{lstlisting}

\paragraph{Evaluation query usage.}
\label{app:workspace_query_usage}

\begin{figure}[t]
    \centering
    \vspace{-8em}
    \includegraphics[width=\linewidth]{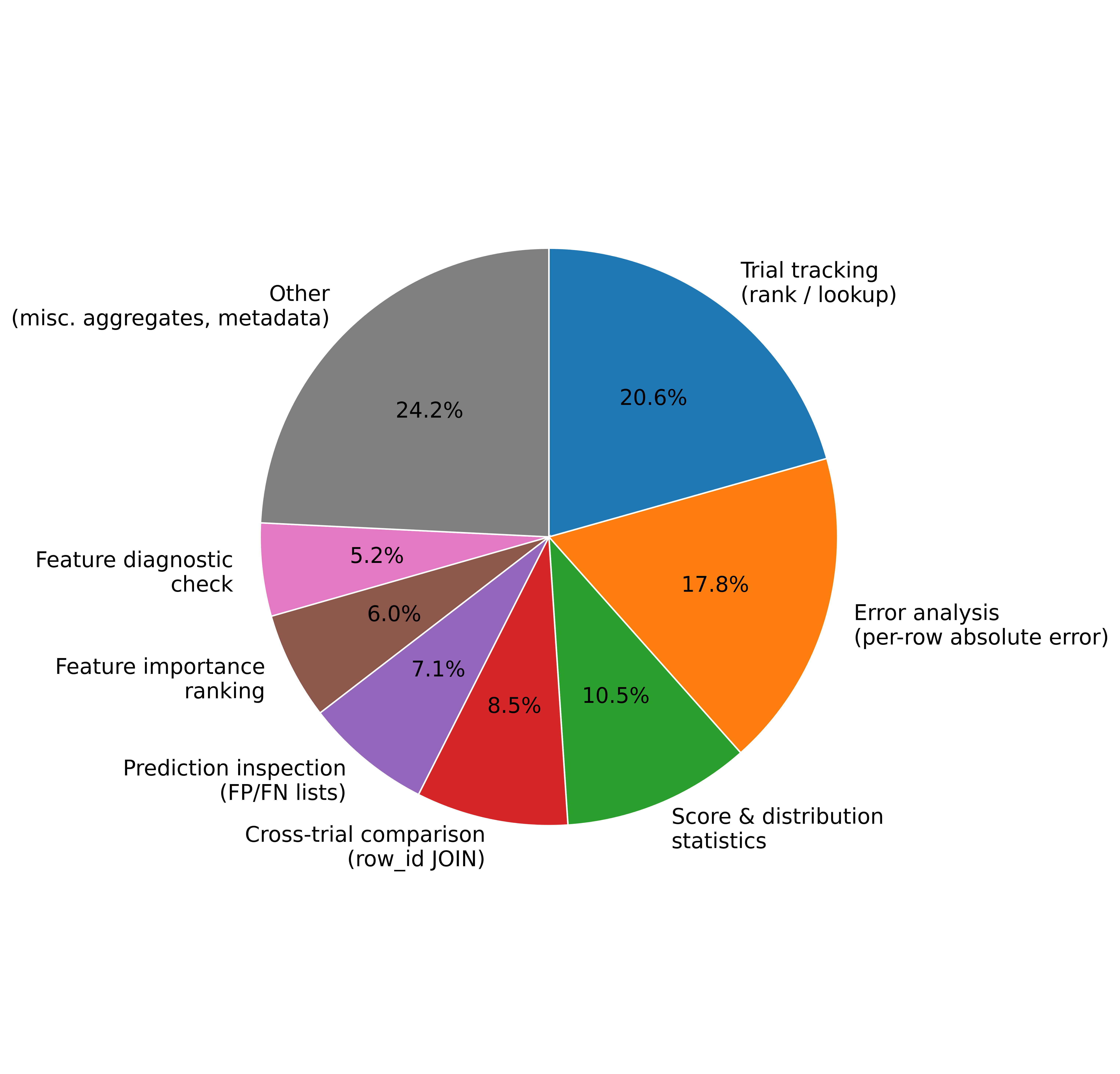}
    \vspace{-8em}
    \caption{Distribution of evaluation workspace queries across all agent runs.}
    \label{fig:workspace_query_breakdown}
\end{figure}

To characterize how the agent uses \texttt{query\_eval\_workspace()} in practice, we analyzed agent runs across all benchmarks.
A majority of runs issue at least one evaluation query, and \Cref{fig:workspace_query_breakdown} summarizes the distribution of these queries by intent.
The dominant patterns reflect an iterative diagnostic strategy:
\begin{enumerate}[label=(\arabic*),itemsep=0pt, topsep=0pt, parsep=0pt, partopsep=0pt]
\item \emph{Trial tracking} (20.6\%) covers queries that rank or retrieve trial records by validation score to monitor search progress.
\item \emph{Error analysis} (17.8\%) computes per-row prediction errors, often sorted to expose the worst-predicted examples.
\item \emph{Score and distribution statistics} (10.5\%) aggregates prediction scores and class-conditional summaries for a single trial.
\item \emph{Cross-trial prediction comparison} (8.5\%) joins two trials on the stable \texttt{row\_id} to identify examples whose predictions changed after a feature revision.
\item \emph{Prediction inspection} (7.1\%) retrieves ranked lists of false-positive or false-negative examples.
\item \emph{Feature importance ranking} (6.0\%) and \emph{feature diagnostic checks} (5.2\%) inspect feature-level signals to identify low-value or potentially defective feature blocks.
\item The remaining queries (24.2\%) cover miscellaneous aggregates, metadata lookups, and other auxiliary diagnostics.
\end{enumerate}

Together, these patterns reveal a structured self-diagnostic loop: the agent first tracks which trial is best, then drills into row-level failures, compares the impact of individual feature changes, and finally uses feature-level evidence to diagnose possible causes and guide the next feature revision.

\paragraph{Anti-leakage constraint.}
Workspace tables are analysis-only: they contain validation predictions and labels, and must not be used by feature queries or predictive models.
The \texttt{query\_eval\_workspace()} interface is a read-only diagnostic interface over the evaluation workspace
\[
\mathcal W^{(\leq b)}
=
\left\{
\widetilde{\mathcal D},
\mathcal T_{\mathrm{trials}}^{(\leq b)},
\mathcal T_{\mathrm{preds}}^{(\leq b)}
\right\}.
\]
Thus, evaluation queries may join validation outcomes back to tables in \(\widetilde{\mathcal D}\) for error analysis, for example, to aggregate errors by entity attributes, timestamps, or relational context.
However, these queries are never used as predictive features. Candidate feature programs submitted to \texttt{validate\_program()} are executed only through the feature-query path, and the validation harness does not expose workspace tables such as \texttt{trials} or \texttt{eval\_predictions} to feature construction during validation or inference.

\paragraph{Point-in-time correctness.}
For each validation or test split, the system exposes the target rows through \texttt{eval\_table}, which contains \texttt{row\_id}, the entity key, and the prediction timestamp. Feature queries are evaluated against this split-specific table and are merged by \texttt{row\_id}. Thus, two target rows with the same entity but different timestamps can receive different feature values. The prompts instruct the agent to filter all timestamped source tables using predicates of the form \texttt{source\_time < eval\_table.timestamp}. In addition, our validation system checks that submitted feature queries are anchored on \texttt{eval\_table} and return \texttt{row\_id}; queries that cannot be aligned row-wise are rejected.

\begin{table}[t]
\centering
\small
\setlength{\tabcolsep}{6pt}
\renewcommand{\arraystretch}{1.10}

\begin{subtable}[t]{0.48\linewidth}
\centering
\begin{tabular}{@{}ll@{}}
\toprule
\textbf{Field} & \textbf{Role} \\
\midrule
\texttt{trial\_id} & primary key \\
\texttt{trial\_name} & trial label \\
\texttt{split} & data split \\
\texttt{model\_choice} & model configuration \\
\texttt{feature\_block\_names} & feature groups \\
\texttt{feature\_query\_hash} & program identifier \\
\texttt{primary\_metric} & validation metric \\
\texttt{primary\_score} & validation score \\
\texttt{metrics\_json} & serialized metrics \\
\bottomrule
\end{tabular}
\caption{\texttt{trials}}
\label{tab:trials_schema}
\end{subtable}
\hfill
\begin{subtable}[t]{0.48\linewidth}
\centering
\begin{tabular}{@{}ll@{}}
\toprule
\textbf{Field} & \textbf{Role} \\
\midrule
\texttt{trial\_id} & foreign key to \texttt{trials} \\
\texttt{row\_id} & stable row index \\
\texttt{entity\_id} & target entity \\
\texttt{label} & ground-truth label \\
\texttt{score} & predicted score \\
\texttt{predicted\_class} & predicted class \\
\texttt{split} & data split \\
\bottomrule
\end{tabular}
\caption{\texttt{eval\_predictions}}
\label{tab:eval_predictions_schema}
\end{subtable}

\caption{
\textbf{Schema of the evaluation workspace.}
The \texttt{trials} table stores trial-level metadata, model configurations, feature-program identifiers, and validation metrics.
The \texttt{eval\_predictions} table stores row-level validation outputs for each trial.
The field \texttt{row\_id} is stable within each split, enabling cross-trial comparisons on the same target examples.
}
\label{tab:eval_workspace_schema}
\end{table}

\section{Case study}\label[appendix]{sec:case_study}
To better understand the search process of \textsc{RelAgent}, we analyze one complete agent run in detail. The goal of this case study is to characterize how the agent structures its search.

We study a run on the \texttt{user-engagement} task using the \texttt{rel-stack} dataset. Stack Exchange is a network of question-and-answer websites organized around different topics. The prediction task is to determine, for a user at a given prediction timestamp, whether that user will make any votes, posts, or comments in the following three months. 

\subsection{Agent search summary}
The analyzed run consists of 45 conversation turns and 30 validation trials. Of these validation trials, 28 produced validation scores, while 2 failed due to execution issues: one timeout and one SQL error. The agent made 16 direct SQL exploration calls over the training database, consisting of one table-listing call, eight schema/table-info calls, and seven exploratory SQL queries. It also made 7 calls to the evaluation workspace, which stores validation histories and row-level predictions from previous trials. 

Across the run, the agent proposed 178 distinct feature columns in successful validation diagnostics and ultimately selected a final champion program with 45 merged features using the XGBoost DART model. The complete conversation trace is provided in the \url{https://anonymous.4open.science/r/RelAgent}.

We manually segmented the agent’s search into four semantic blocks. This segmentation is not used by the agent itself; rather, it is an analysis of the trace intended to expose the structure of the search. 

\begin{itemize}[leftmargin=0.5cm, itemsep=2pt, topsep=0pt]
    \item \textbf{Initial database exploration and baseline programs (Turn 0-1).}
    The agent first inspects the database structure. It lists the available tables, including \texttt{badges} \footnote{\textit{Badges} on Stack Exchange are achievement-based indicators awarded to users for specific contributions and participation milestones within the platform’s community.}, \texttt{comments}, \texttt{post history}, \texttt{post links}, \texttt{posts}, \texttt{users}, \texttt{votes}, and the training table, and then queries schemas and row counts. It observes that the classification problem is strongly imbalanced. The first submitted program uses LightGBM with standard gradient boosting and contains three main feature groups: user-profile features, activity counts over posts, comments, and votes, and recency features measuring time since previous activity. This first validation reaches about 71.72 AUROC on the validation set. A second, similar validation gives nearly the same score.
    \item \textbf{Revision of vote and activity features using validation feedback. (Turn 2-6).} After the initial baselines, the agent consults previous trial results in the evaluation workspace and notices that vote-related features, such as upvotes, downvotes, and accepted-vote counts, were either constant or had low SHAP value. The agent removes these columns and replaces them with voting-\textit{breadth} features, counting the number of \textit{distinct} posts a user voted on within recent windows. This produces the first clear performance increase, with validation AUROC rising to about 78.20. At this point, the feature program contains the four semantic blocks that remain the core for most later trials: recency, activity windows, comments and votes, and user profile. The agent then tries additional aggregates, such as active-day counts and velocity ratios, but these expanded variants reduce validation performance and the agent decides to focus on model changes instead.
    \item \textbf{Model-family search with the core SQL fixed. (Turn 7-28).} In this phase the agent explores variations of the predictive model while keeping the feature set mostly fixed. It first evaluates a DART variant of LightGBM and improves the validation score to about 78.80. Other LightGBM configurations, including GOSS, stronger regularization, smaller tree depth, and different learning rates, do not improve over this DART configuration. The agent then switches to \textit{XGBoost}-DART, again using the same four core SQL blocks. This produces a large jump in validation AUROC to approximately 89.90. It then validates nearby XGBoost-style configurations, varying depth, learning rate, and child-weight parameters, and also compares against plain XGBoost and CatBoost. These trials remain close to the XGBoost-DART score but do not exceed the first XGBoost-DART trial. 
    \item \textbf{Incremental feature expansion on top of the fixed core (Turn 29-44).}
    In the final part of the run, the agent keeps XGBoost-DART and the four-query core fixed while adding feature blocks from additional tables: It adds badge-based features, including recent badge counts, and then adds post-history features based on user-related events in the post-history table. Both additions slightly improve validation performance, reaching about 90.30 AUROC. The agent also tests post-link features, but these lower the score and are not retained. The badge features are then refined with badge-class splits, days since last badge, and badge-velocity ratios. A richer post-history variant is also tried but later reverted after a small performance drop. Trial 29, which combines the four core blocks with badge and post-history is selected as the final champion.
\end{itemize}
An overview of the number of tool calls per turn as well as the running best validation AUROC is provided in Figure \ref{fig:case_study}b).

\subsection{Target features found by the agent}

\begin{figure}[h]
    \centering
    \begin{subfigure}{0.35\textwidth}
        \centering
        \includegraphics[width=\linewidth]{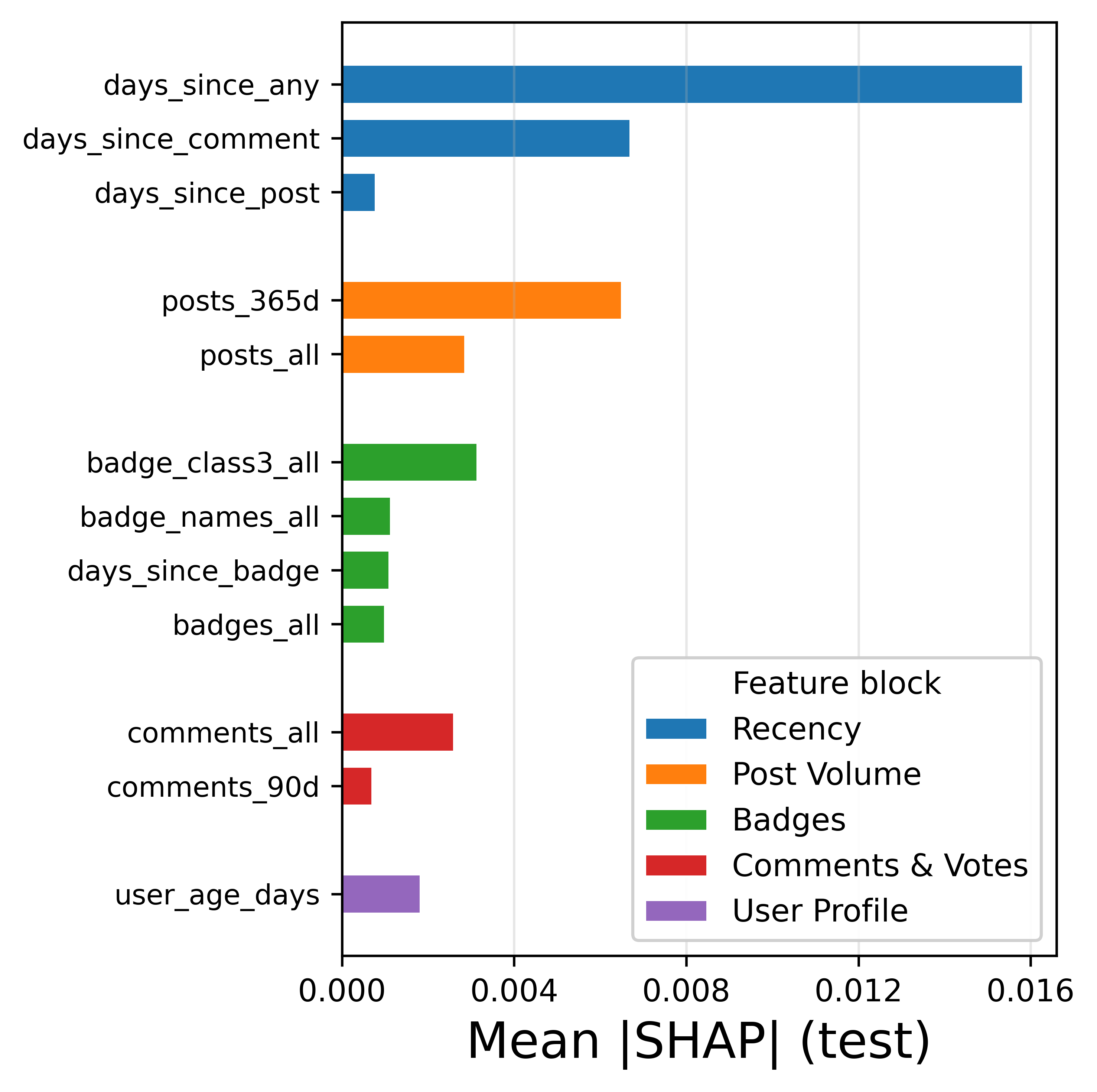}
        \caption{\textbf{Feature-block importance}}
    \end{subfigure}
    \hfill
    \begin{subfigure}{0.5\textwidth}
        \centering
        \includegraphics[width=\linewidth]{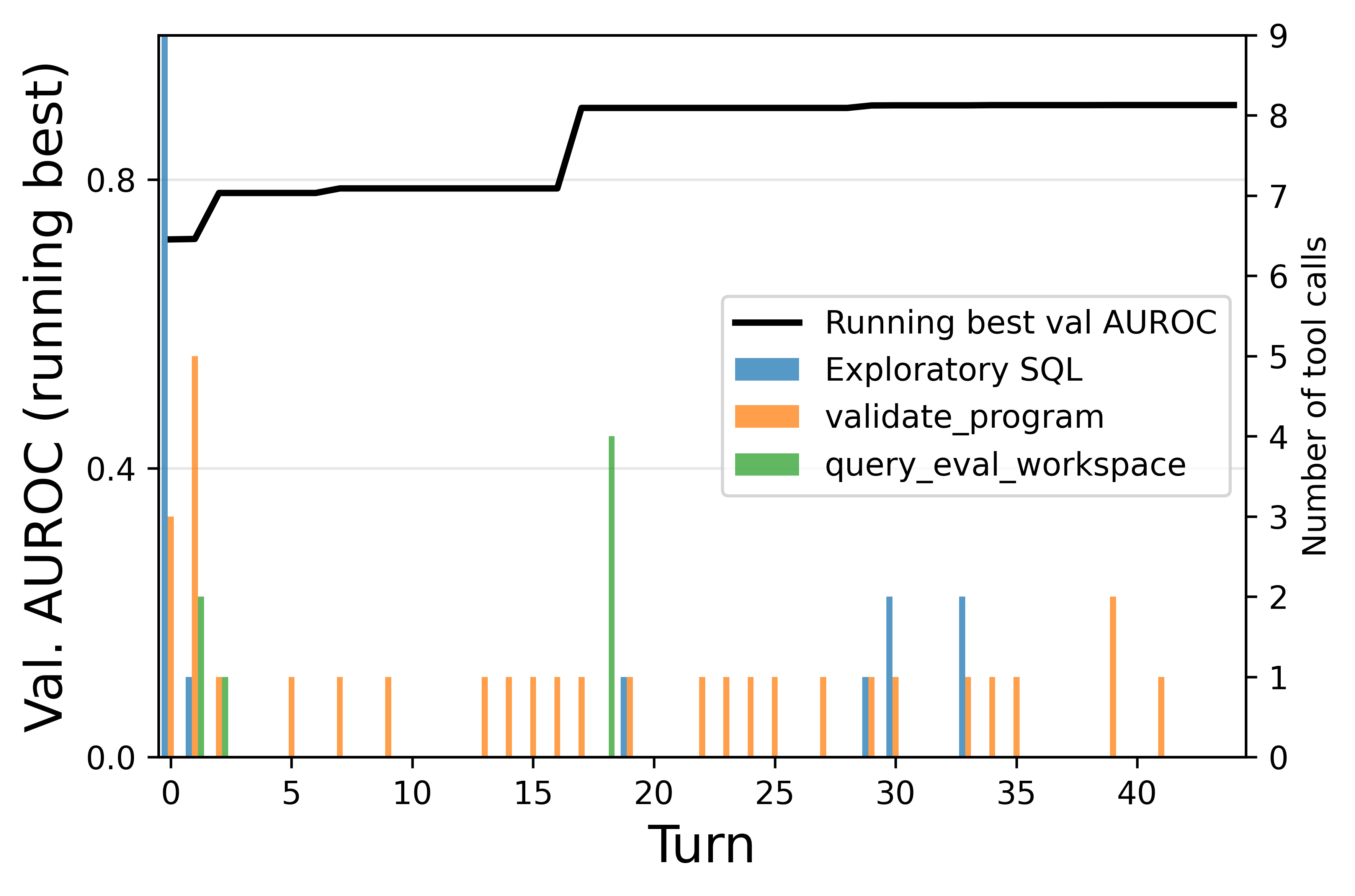}
        \caption{\textbf{Search trajectory}}
    \end{subfigure}
    \caption{
    Case-study summary for the inspected \texttt{rel-stack}/\texttt{user-engagement} run.
    Figure (a) shows the mean absolute SHAP values of the most influential test-set features, grouped by the feature block from which they originate.
    Figure (b) shows the running best validation AUROC over conversation turns together with the number of tool calls made in each turn.
    }
    \label{fig:case_study}
\end{figure}

Since the final predictor is based on explicit SQL features, its representation can be inspected directly. Note that a single SQL query can construct multiple feature columns. We refer to all features originating from the same query as one feature \textit{block}. The final program in the inspected run consists of seven such blocks. Figure~\ref{fig:case_study}a reports the mean absolute SHAP values of all features with mean absolute SHAP value greater than $0.0005$, grouped by feature block. Features within a block are typically semantically related.

The most important block, \textbf{Recency}, measures how recently the user was active before the prediction time. It includes features such as the number of days since the user's last post, comment, or vote. A second block, \textbf{Post Volume}, captures how much the user has posted historically and in recent rolling windows, including lifetime question and answer counts as well as windowed post activity. The \textbf{Badges} block represents badge acquisition behavior, including badge counts, distinct badge names, class-specific badge counts, recency of the last badge, and short-versus-long window velocity ratios. The \textbf{Comments and Votes} block measures comment and voting behavior in rolling windows, including the number of distinct posts voted on. This block emerged from the agent's correction of constant vote-type features during the search. Finally, the \textbf{User Profile} block contains basic user-level information, primarily account age and whether the user appears in the user table.

\section{Synthetic experiments on directed triangle detection}
\label[appendix]{app:synthetic_triangle}

To evaluate whether the agent can discover a precise relational pattern from labeled examples alone without any schema hints about the underlying rule, we construct a controlled synthetic benchmark based on directed triangle membership. 

\paragraph{Setup.}
Each dataset consists of a single directed edge relation \texttt{R(src, dst)} drawn from an Erdős--Rényi random graph ($n=300$ nodes, edge probability $p=0.02$). 
Training uses 5 independent graphs (1500 nodes total); validation and test each use 2 graphs (600 nodes). Node IDs are globally unique across graphs, so the agent must learn a \emph{transferable} structural rule rather than memorize node identities. The agent can write SQL feature queries over relational table \texttt{R}, and a LightGBM classifier is trained automatically on those features. The agent observes validation AUROC after each trial and iteratively refines its queries up to a budget of 30 trials, the same as the setup in the main experiments.  

\begin{figure}[t]
\centering

\begin{minipage}[t]{0.45\textwidth}
\centering
\small\textbf{Task 1}: no self-loops

\[
\begin{tikzcd}[row sep=large, column sep=large]
|[draw,circle,fill=green!20]| a \arrow[r] &
|[draw,circle,fill=green!20]| b \arrow[dl] \arrow[d] \\
|[draw,circle,fill=green!20]| c \arrow[u] &
|[draw,circle,fill=gray!20]| d
\end{tikzcd}
\]

\end{minipage}
\hfill
\begin{minipage}[t]{0.45\textwidth}
\centering
\small\textbf{Task 2}: with self-loops

\[
\begin{tikzcd}[row sep=large, column sep=large]
|[draw,circle,fill=green!20]| a
  \arrow[r]
  \arrow[loop left]
&
|[draw,circle,fill=green!20]| b
  \arrow[d]
  \arrow[dl]
  \arrow[loop right]
\\
|[draw,circle,fill=green!20]| c
  \arrow[u]
  \arrow[loop left]
&
|[draw,circle,fill=gray!20]| d
  \arrow[loop right]
\end{tikzcd}
\]

\end{minipage}
\vspace{1em}
\caption{
\textbf{Synthetic directed triangle tasks.} Green nodes indicate positives (satisfying $Q(x)$) and gray nodes indicate negatives (not satisfying $Q(x)$).
Left: loop-free setting. Right: self-loop setting.
}
\label{fig:synthetic_triangle_tasks}
\end{figure}
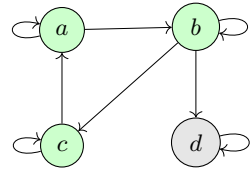

We distinguish two closely related predicates. The first is the directed
3-walk closure predicate
\[
  P(x) =
  \exists y,z:\;
  \texttt{R}(x,y) \wedge \texttt{R}(y,z) \wedge \texttt{R}(z,x),
\]
which can be computed by a three-way self-join. The second is the \emph{proper} directed 3-cycle predicate
\[
  Q(x) =
  \exists y,z:\;
  \texttt{R}(x,y) \wedge \texttt{R}(y,z) \wedge \texttt{R}(z,x)
  \wedge x \neq y \wedge y \neq z \wedge z \neq x .
\]
The two predicates coincide on loop-free graphs, but differ when self-loops are present. As illustrated in \Cref{fig:synthetic_triangle_tasks}, the loop-free task and the self-loop task differ in whether degenerate directed walks are possible.
In Task~1, positives are nodes participating in a directed 3-cycle, and the
absence of self-loops makes the distinctness constraint implicit. In Task~2,
self-loops introduce spurious matches for the naive three-edge walk, so the
agent must distinguish proper directed cycles from degenerate self-loop walks.

Note that both of these predicates $P(x),Q(x)$ are not expressible by standard message-passing graph neural networks, since it is not definable in graded modal logic.

\paragraph{Task 1: directed triangle detection on loop-free graphs.}

In the first task, all self-loops are removed from \texttt{R} (by setting $p_{\mathrm{sl}} = 0.0$), and the binary label for node \(x\) is given by \(Q(x)\). Since no edge of the form \(\texttt{R}(u,u)\) exists, any satisfying assignment to \(P(x)\) must already use three distinct nodes. Therefore, on this task,
\[
  P(x) \equiv Q(x).
\]
Consequently, the agent only needs to discover the directed 3-cycle pattern: the raw three-way self-join for \(P(x)\) is already an exact feature for the target. This task tests whether the agent can find the relevant relational motif from examples, but it does not require the distinctness constraints to be identified separately.

\paragraph{Task 2: directed triangle detection on graphs with self-loops.}
The second task keeps the same target label \(Q(x)\), but adds self-loops to the graph. Each node \(u\) receives a self-edge \(\texttt{R}(u,u)\) independently with probability \(p_{\mathrm{sl}}\). In this setting, \(P(x)\) and \(Q(x)\) are no longer equivalent. In particular, when \(p_{\mathrm{sl}}=1.0\), every node satisfies \(P(x)\) through the degenerate walk
\[
  x \to x \to x \to x,
\]
even if it does not participate in any proper directed 3-cycle. Thus, the naive three-way self-join over-predicts positives and thus loses discriminative power.
Consequently, solving Task 2 requires the agent to discover the proper-cycle condition \(Q(x)\), either by explicitly enforcing the inequalities \(x \neq y\), \(y \neq z\), and \(z \neq x\), or equivalently by filtering self-loop edges before performing the directed 3-cycle join, e.g., \texttt{WHERE src != dst}.

\begin{table}[t]
\centering
\small
\caption{\textbf{Directed triangle detection on synthetic graphs} ($p_{\mathrm{sl}}$: self-loop probability). Each row is one independent agent run. ``Discovery trial'' is the trial at which the winning query was first proposed; ``---'' indicates the agent never reached AUROC~1.0.}
\begin{tabular}{lccl}
\toprule
\textbf{Setup} & \textbf{$p_{\mathrm{sl}}$} & \textbf{Discovery trial} & \textbf{Agent strategy}\\
\midrule
No self-loops      & 0.0 & 1   & Raw \texttt{cycle3\_cnt}; no inequality constraint required \\
\midrule
Partial self-loops & 0.4 & 1   & \texttt{WHERE src $\neq$ dst} written immediately \\
Partial self-loops & 0.4 & --- & Proxy features only; inequality constraint never written \\
\midrule
Hard self-loops    & 1.0 & 14  & Filter + directed join combined at trial 14 \\
Hard self-loops    & 1.0 & --- & Raw \texttt{cycle3\_cnt + cycle2\_cnt}; no inequality constraint \\
Hard self-loops    & 1.0 & --- & \texttt{cycle3\_out + recip\_out}; no inequality constraint \\
\bottomrule
\end{tabular}
\label{tab:synthetic_triangle}
\end{table}
\begin{figure}[h]
    \centering
    \includegraphics[width=0.7\linewidth]{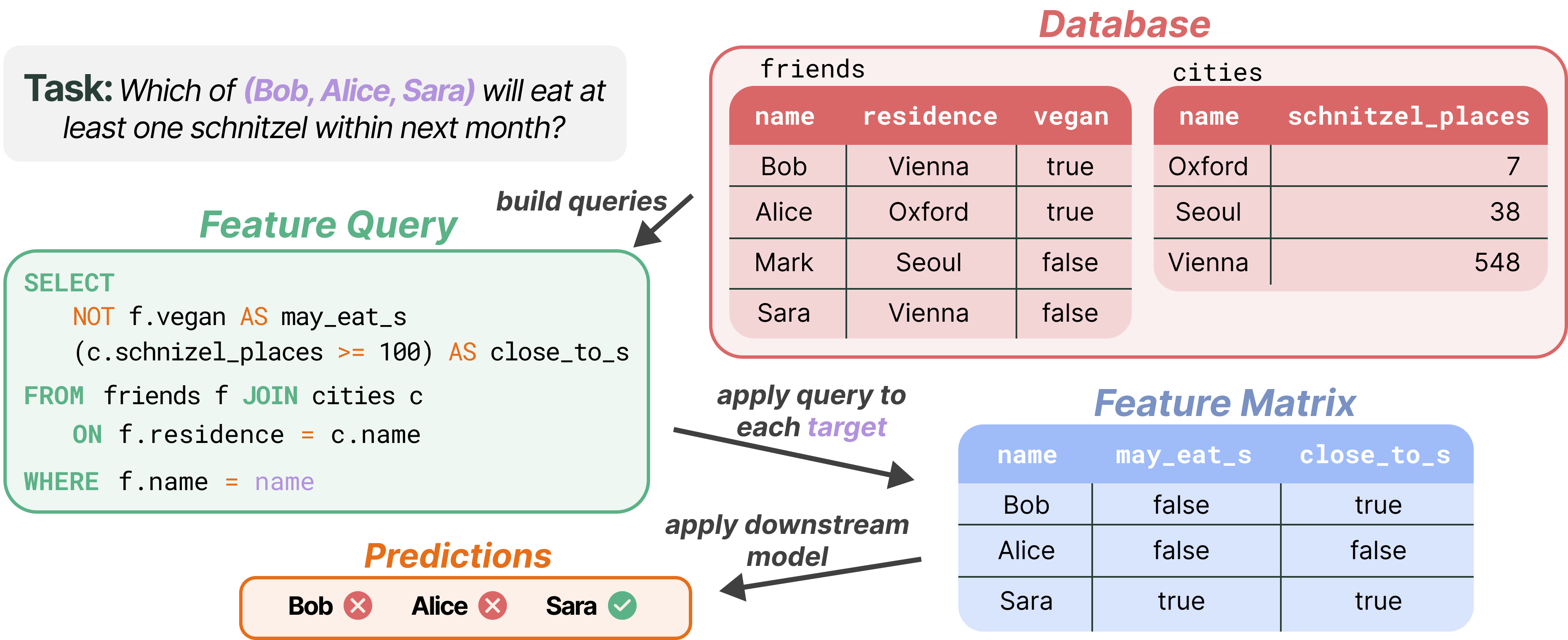}
    \caption{Illustrative example of SQL-based feature construction over relational data. 
    }
\label{fig:feature_construction_example}
\end{figure}

\paragraph{Results.}
\Cref{tab:synthetic_triangle} summarises representative run per setup.
On Task 1 (no self-loop), the agent reliably solves the task on the first trial, since the standard 3-way self-join is sufficient and LightGBM learns the threshold automatically.
Task~2 (with self-loop) separates the two predicates. With partial self-loops
(\(p_{\mathrm{sl}}=0.4\)), the agent sometimes discovers the
\texttt{src != dst} filter immediately, recovering the proper-cycle signal.
In other runs, it relies on proxy features that correlate with \(Q(x)\) but do not exactly encode the target. With self-loops on every node (\(p_{\mathrm{sl}}=1.0\)), the distinction becomes necessary: the unfiltered predicate \(P(x)\) fires for all nodes through degenerate self-loop walks. In one of three independent runs, the agent reaches AUROC 1.0 by first discovering self-loop filtering in an imperfect feature and later combining it with the directed 3-cycle join at trial 14. The winning query is:

\begin{lstlisting}[language=SQL, basicstyle=\ttfamily\footnotesize, frame=single]
WITH edges AS (SELECT src, dst FROM R WHERE src != dst),
cycle3 AS (
  SELECT e1.src AS node_id, COUNT(*) AS cycle3_cnt
  FROM edges e1
  JOIN edges e2 ON e2.src = e1.dst
  JOIN edges e3 ON e3.src = e2.dst AND e3.dst = e1.src
  GROUP BY e1.src
)
SELECT e.node_id, COALESCE(c.cycle3_cnt, 0) AS cycle3_cnt
FROM eval_table e LEFT JOIN cycle3 c USING(node_id)
\end{lstlisting}

The two failing runs never wrote \texttt{src $\neq$ dst}; they relied on raw (inflated) cycle counts, which provide a strong but imperfect signal because every node with a self-loop accumulates at least one spurious 3-cycle path ($u \to u \to u \to u$), inflating its count regardless of true triangle membership.
Task~1 shows that the agent can discover the directed 3-cycle pattern when \(P\) and \(Q\) coincide; Task~2 shows that self-loop variants make the distinction between \(P\) and \(Q\) observable, and therefore test whether the agent can recover the additional first-order distinctness constraint from validation feedback.

\section{Deep feature synthesis (DFS) vs. agentic feature construction}
\label[appendix]{app:dfs}

\subsection{Expressivity}\label{sec:dfs-expressivity}
\textsc{RelAgent} constructs a task-dependent feature map from a relational database to a feature matrix by synthesizing SQL feature queries with an LLM agent and applying a downstream predictor (see \Cref{fig:feature_construction_example} as an example). This invites a comparison to RDBLearn~\citep{xu2026rdblearn}, which also adopts a two-stage pipeline by first summarizing relational data into a single table and then applying a tabular foundation model~\citep{tabpfnv2}. However, in contrast to \textsc{RelAgent}, this mapping from relational database to a single table is not task-dependent but instead relies on a fixed, task-agnostic procedure, leading to a gap in expressivity.

Specifically, the deep feature synthesis (DFS)~\citep{kanter2015deep} used in \citet{xu2026rdblearn} constructs features via predefined, column-wise aggregations. As first observed by \citet{hudovernik2026kumorfm} such operations are inherently limited, as they cannot capture interactions \textit{across rows}.

To illustrate this limitation, consider a database with a target table $\mathcal{T}_1$ and a child table $\mathcal{T}_2$ containing binary features $A$ and $B$. For each entity in $\mathcal{T}_1$, define the target $y = 1$ if and only if $A$ and $B$ co-occur in at least one related row of $\mathcal{T}_2$. Any model operating on column-wise aggregations will fail to correctly distinguish these cases, since the joint occurrence information is not preserved.

In contrast, \textsc{RelAgent} can directly express this target function via the following SQL query:
\begin{lstlisting}[language=SQL, basicstyle=\ttfamily\footnotesize, frame=single]
SELECT SUM(CASE WHEN A = 1 AND B = 1 THEN 1 ELSE 0 END) > 0 AS y
FROM T2
WHERE t1_id = entity
\end{lstlisting}

\subsection{Interpretability}
Each feature used by \textsc{RelAgent} is defined by an explicit SQL query. Predictions can therefore be inspected through the selected queries and the downstream model’s feature usage (see also the case study in Section~\ref{fig:case_study}). This yields a transparent pipeline in which both feature construction and prediction remain accessible and interpretable.

Importantly, through the evaluation workspace, \textsc{RelAgent} can iteratively assess candidate features and retain only those that are most informative for the task. As a result, the final feature set is compact and focused on relevant signals, rather than relying on a large collection of generic aggregations.

\begin{figure}[t]
    \centering
    \includegraphics[width=0.8\textwidth]{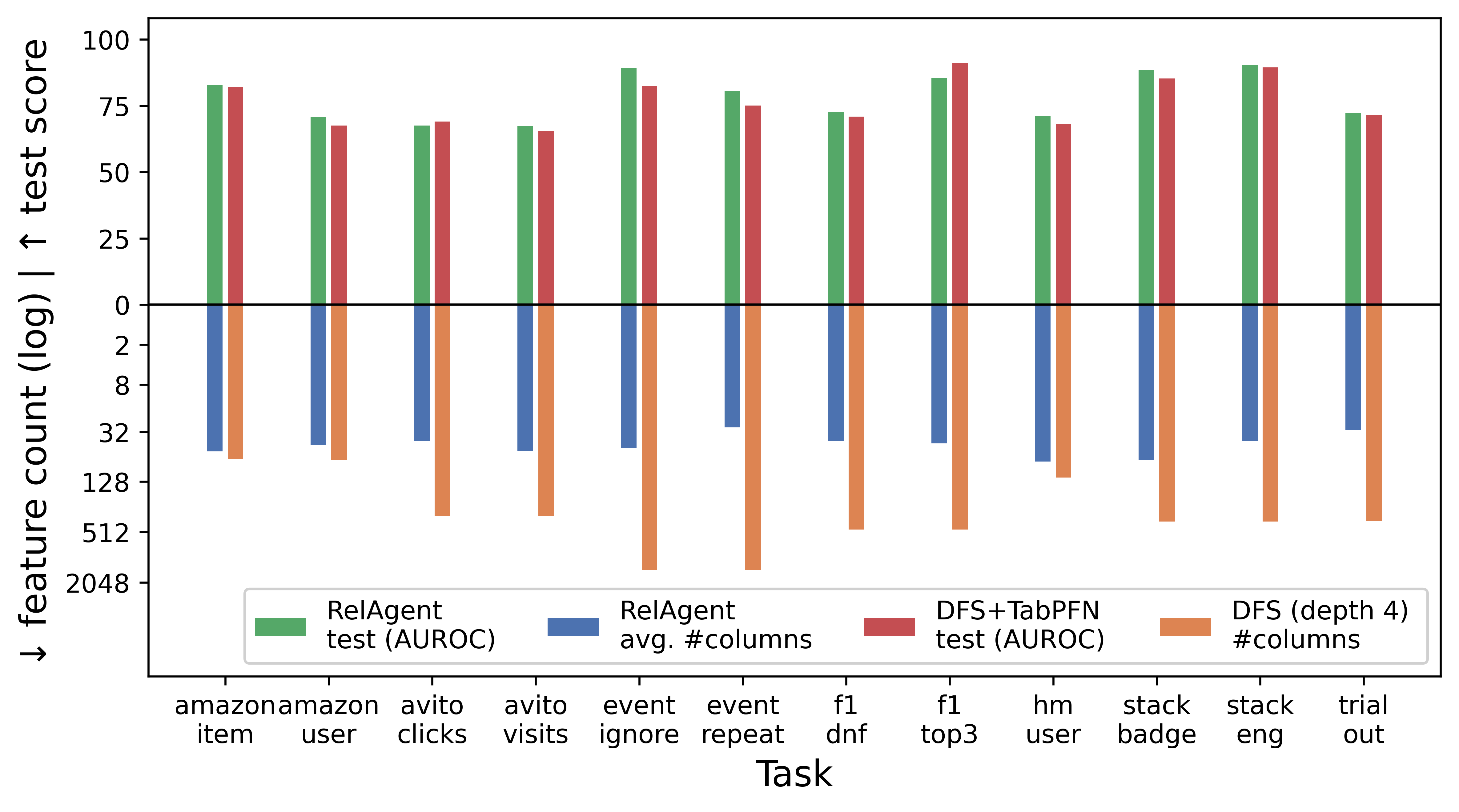}
    \caption{Test scores (AUROC) on RelBenchV1 classification tasks of \textsc{RelAgent} and DFS+TabPFN~\cite{kanter2015deep,xu2026rdblearn} (↑) as well as the \textbf{number of feature columns produced per task} by \textsc{RelAgent} and DFS (depth 4) (↓).}
    \label{fig:dfs}
\end{figure}

In contrast, DFS-based approaches rely on predefined aggregation primitives applied in a breadth-first manner over the relational schema. While this provides broad coverage, it often leads to a large number of features. As shown in Figure~\ref{fig:dfs}, \textsc{RelAgent} achieves competitive to superior performance while producing significantly fewer features (note the log-scale in Figure~\ref{fig:dfs}). Due to the expansion mechanism of DFS, the number of generated features grows with the degree of the target table and its related tables. In strongly connected relational databases, this can result in a substantial increase in feature columns (see the \texttt{rel-event} dataset in Figure~\ref{fig:dfs}). In contrast, the number of features created by \textsc{RelAgent} stays constant over different tasks and datasets.

\section{Search budget and convergence}
\label[appendix]{app:search_budget}

\begin{figure}[t]
    \centering
    \includegraphics[width=0.8\linewidth]{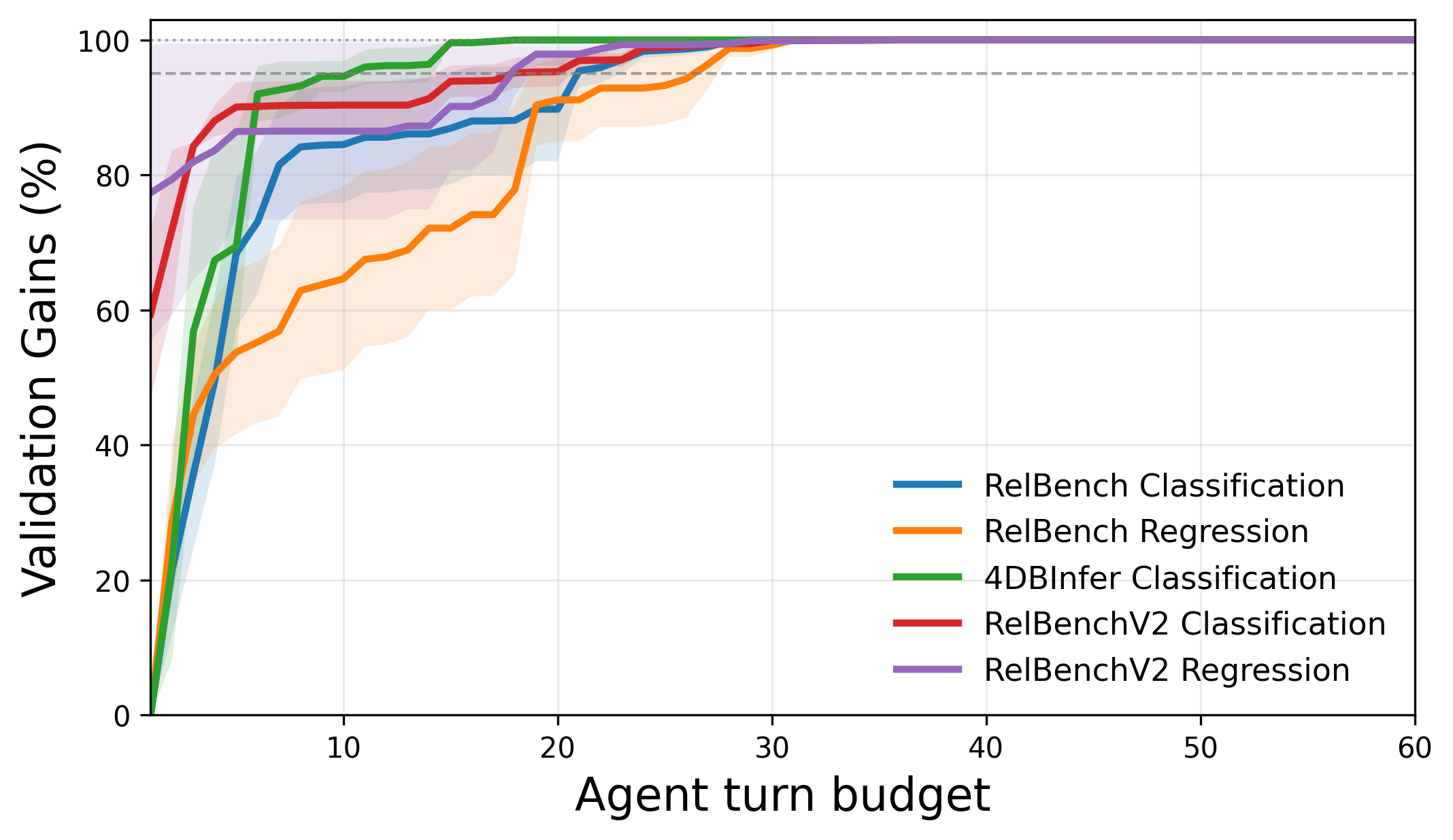}
    \caption{\textbf{Search-budget scaling.} The fraction of final validation gain recovered as a function of agent turn budget. Shaded regions indicate variation across tasks.}
    \label{fig:search_budget}
\end{figure}

\noindent\textbf{Setup.} We study how \textsc{RelAgent} scales with the number of allowed agent turns. For each task, we compute the best validation score achieved up to turn \(k\), and normalize the improvement by the final best validation score reached within the full 60-turn budget. Thus, a value of \(100\%\) indicates that the best final validation score has already been recovered by turn \(k\).

\noindent\textbf{Results.} As shown in \Cref{fig:search_budget}, most gains occur early in the search. On classification tasks, \textsc{RelAgent} often recovers a large fraction of its final validation gain within the first 5--10 turns, especially on 4DBInfer and RelBenchV2 classification. Regression tasks converge more gradually, reflecting the larger role of objective choice, target transformations, and model configuration in addition to feature discovery. Across all benchmark groups, performance largely saturates before the full 60-turn budget, suggesting that the agent benefits from iterative feedback but does not require long unbounded search in most cases.

\section{Efficiency analysis}
\label[appendix]{app:efficiency}

We analyze the wall-clock time and token consumption of our agent across three backbone LLMs: GPT-5.2, Qwen3-32B, and DeepSeek-V4-Pro.
For each completed run, we parse the agent's conversation trace and count tokens.  Wall-clock time is measured from the first to the last event in the trace.  Thinking tokens, i.e., the content within \texttt{<think>} \dots \texttt{</think>} delimiters, are reported separately because they consume inference compute but are never passed back to the agent as context.

\paragraph{Setup.}
Each run consists of up to 60 agent turns over a single RelBench task.
At each turn, the model receives the full conversation history and produces a response that may include zero or more tool invocations (SQL queries, \texttt{validate\_program}, etc.); all tool results are collected and appended before the next turn begins.
Since a single turn can issue multiple tool calls, the total tool invocations per run can exceed the number of turns.
All three models are queried through the same internal API gateway with temperature~$= 1$ and a maximum of 8\,000 output tokens per turn. GPT-5.2 and DeepSeek-V4-Pro are accessed via LiteLLM API calls through an internal API gateway.  Qwen3-32B is served locally using vLLM on 2$\times$ NVIDIA RTX~6000~Ada (48\,GB each, tensor-parallel).  All models use temperature~$= 1$ and a maximum of 8\,000 output tokens per turn.

\begin{table}[t]
\centering
\caption{%
  \textbf{Efficiency of three backbone LLMs} per agent run (mean $\pm$ std) averaged over RelBench V1 classification and regression runs. Thinking tokens are \texttt{<think>} chain-of-thought
  content generated by Qwen3-32B, not re-injected as context.  
}
\label{tab:efficiency}
\setlength{\tabcolsep}{6pt}
\begin{tabular}{lrrr}
\toprule
& \textbf{GPT-5.2} & \textbf{Qwen3-32B} & \textbf{DeepSeek-V4-Pro} \\
\midrule
Wall-clock time (min)   & $35.2 \pm 27.8$  & $97.5 \pm 63.9$  & $44.5 \pm 49.1$ \\
Agent turns per run     & $41.1 \pm 11.7$  & $44.2 \pm 15.2$  & $40.7 \pm 31.0$ \\
Tool invocations / run  & $47.2 \pm 10.9$  & $25.2 \pm 6.7$   & $31.0 \pm 16.1$ \\
\midrule
Input tokens            & $22{,}270 \pm 7{,}356$  & $6{,}343 \pm 2{,}509$   & $8{,}404 \pm 3{,}600$ \\
Output tokens           & $20{,}671 \pm 5{,}884$  & $31{,}873 \pm 11{,}118$ & $26{,}744 \pm 35{,}034$ \\
Thinking tokens         & ---                     & $36{,}782 \pm 10{,}870$ & --- \\
Total tokens  & $42{,}941 \pm 11{,}911$ & $74{,}998 \pm 22{,}267$ & $35{,}148 \pm 38{,}634$ \\
\midrule
\textbf{Trace-token cost / run (USD)} & $\mathbf{\approx\$1.57}$ & $\mathbf{\$0}$ & $\mathbf{\approx\$0.11}$ \\
\bottomrule
\end{tabular}
\end{table}

\paragraph{Time efficiency, token efficiency, and cost.}
GPT-5.2 and DeepSeek-V4-Pro complete a run in 35 and 45 minutes on average, respectively.  Qwen3-32B is $2.8\times$ slower (98~min) due to the latency of its extended chain-of-thought generation over two locally-hosted RTX~6000 GPUs.

Input tokens are counted from the conversation trace using the \texttt{cl100k\_base} tokenizer (system prompt counted once plus all tool results). GPT-5.2 has substantially more input tokens ($\approx$22K) than Qwen3-32B and DeepSeek-V4-Pro ($\approx$6--8K) as it issues roughly twice as many tool calls per run, accumulating more SQL query results in its input stream.

In terms of total tokens, Qwen3-32B generates $1.7\times$ more tokens than GPT-5.2; 49\% of Qwen3-32B's tokens are hidden thinking tokens never re-injected as context, so the visible output (32K) is only
$1.5\times$ larger than GPT-5.2's (21K).

In terms of cost, DeepSeek-V4-Pro costs~\citep{deepseek-v4-pro-cost} only \$0.11 vs.\ \$1.57 for GPT-5.2~\citep{gpt_5.2_cost}, which is $14\times$ cheaper thanks to DeepSeek's lower per-token price (\$1.74/\$3.48 vs.\ \$15/\$60 per 1M input/output tokens). Qwen3-32B costs \$0 as a locally-hosted model. Costs are estimated from the final conversation trace under an ideal 100\% prompt-cache hit rate: each message is counted once rather than re-counting the full prompt at every API turn. 

\begin{table}[t]
\centering
\caption{\textbf{Effect of the search-turn budget.} Average test AUROC on RelBenchV1 classification tasks. Candidate selection is based on validation performance.}
\label{tab:turn_budget_test}
\small
\begin{tabular}{lcccc}
\toprule
\textbf{Turns} & 10 & 20 & 30 & 60 \\
\midrule
Test AUROC & 77.81 & 77.48 & \textbf{78.87} & 78.38 \\
\bottomrule
\end{tabular}
\end{table}

\begin{table}[t]
\centering
\caption{\textbf{Best-of-$R$ rollout analysis.}
Incremental change in average test AUROC when increasing
the number of independent rollouts from $R-1$ to $R$.
Candidates are selected by validation performance.}
\label{tab:best_of_r}
\small
\setlength{\tabcolsep}{6pt}
\begin{tabular}{lcccc}
\toprule
Benchmark & $+R_2$ & $+R_3$ & $+R_4$ & $+R_5$ \\
\midrule
RelBenchV1 & +0.37 & +0.67 & +0.00 & +0.24 \\
RelBenchV2 & +0.17 & +0.00 & +0.00 & +0.01 \\
4DBInfer & +0.56 & +0.03 & +1.77 & +0.48 \\
\bottomrule
\end{tabular}
\end{table}

\noindent\textbf{Diminishing returns.}
\textsc{RelAgent} achieves near-full-budget performance within ten search turns, recovering approximately 80\% and 90\% of the final validation improvement within 10 and 20 turns, respectively. We further examine the effect of independent rollouts through best-of-$R$ selection, where the best candidate among the first $R$ rollouts is selected by validation performance. Increasing the number of rollouts from one to five yields incremental test AUROC improvements of $(0.37, 0.67, 0.00, 0.24)$ on RelBenchV1, $(0.17, 0.00, 0.00, 0.01)$ on RelBenchV2, and $(0.56, 0.03, 1.77, 0.48)$ on 4DBInfer. While RelBenchV1 and RelBenchV2 exhibit diminishing marginal gains, 4DBInfer benefits substantially from the fourth rollout. Overall, these results suggest that moderate search budgets already capture much of the available performance, although additional exploration can occasionally discover substantially better programs.

\section{Experimental details}
\label[appendix]{app:exp_details}

\begin{table}[t]
\centering
\small
\caption{Database summary and statistics across RelBenchV1, RelBenchV2, and 4DBInfer.}
\setlength{\tabcolsep}{3pt}
\label{tab:dataset_stats}
\begin{tabular}{ll l l r r r r}
\toprule
 & \textbf{Dataset} & \textbf{Domain} & \textbf{Tasks} & \textbf{\#Tasks} & \textbf{\#Tables} & \textbf{\#Rows} & \textbf{\#Cols}  \\
\midrule

\multicolumn{8}{l}{\textbf{RelBenchV1}} \\
\cmidrule(lr){1-8}
 & \texttt{f1}      & Sports      & Race outcome              & 3 & 9  & 74,063       & 67  \\
 & avito   & E-commerce  & User engagement, CTR     & 3 & 8  & 20,679,117   & 42   \\
 & event   & Social      & Event participation      & 3 & 5  & 41,328,337   & 128   \\
 & trial   & Medical     & Clinical trial outcome   & 3 & 15 & 5,434,924    & 140   \\
 & amazon  & E-commerce  & Churn, LTV               & 4 & 3  & 15,000,713   & 15   \\
 & stack   & Social      & Community activities     & 3 & 7  & 4,247,264    & 52  \\
 & hm      & E-commerce  & Churn, LTV               & 2 & 3  & 16,664,809   & 37   \\
\midrule
\multicolumn{8}{l}{\textbf{RelBenchV2}} \\
\cmidrule(lr){1-8}
 & ratebeer   & E-commerce  & Engagement, Churn      & 4 & 13  &  13,787,005    & 221      \\
 & arxiv & Academic Citations & Publications & 2 & 6 & 2,146,112 & 21         \\
 \midrule
\multicolumn{8}{l}{\textbf{4DBInfer}} \\
\cmidrule(lr){1-8}
 & AB   & E-commerce  & Churn                    & 1 & 3  & 24,291,489    & 15      \\
 & OB   & Social      & CTR                      & 1 & 8  & 2,170,441,217 & 31         \\
 & RR   & E-commerce  & CVR                      & 1 & 3  & 23,033,676    & 11          \\
 & SE   & Social      & Churn, Popularity        & 2 & 7  & 5,399,818     & 49  \\

\bottomrule
\end{tabular}
\end{table}

\begin{table}
\centering
\caption{\textbf{Full Ablation results on the entity classification tasks in RelBenchV1.} Higher is better (AUROC). }
\label{tab:ablation-rlv-full}
\setlength{\tabcolsep}{1.5pt}
\footnotesize
\begin{tabular}{clrrrrrrrrrrrr|r}
\toprule
& \multirow{2}{*}{\textbf{Model Variants}} & \multicolumn{2}{c}{\texttt{\small f1}} & \multicolumn{2}{c}{\texttt{\small avito}} & \multicolumn{2}{c}{\texttt{\small event}} & \multicolumn{1}{c}{\texttt{\small trial}} & \multicolumn{2}{c}{\texttt{\small amazon}} & \multicolumn{2}{c}{\texttt{\small stack}} & \multicolumn{1}{c}{\texttt{\small hm}} & \multicolumn{1}{c}{\textbf{\small Avg}}  \\
& & \multicolumn{1}{c}{\texttt{\scriptsize dnf}} & \multicolumn{1}{c}{\texttt{\scriptsize top}} & \multicolumn{1}{c}{\texttt{\scriptsize click}} & \multicolumn{1}{c}{\texttt{\scriptsize visit}} & \multicolumn{1}{c}{\texttt{\scriptsize repeat}} & \multicolumn{1}{c}{\texttt{\scriptsize ignore}} & \multicolumn{1}{c}{\texttt{\scriptsize out}} & \multicolumn{1}{c}{\texttt{\scriptsize user}} & \multicolumn{1}{c}{\texttt{\scriptsize item}} & \multicolumn{1}{c}{\texttt{\scriptsize eng}} & \multicolumn{1}{c}{\texttt{\scriptsize badge}} & \multicolumn{1}{c}{\texttt{\scriptsize churn}} & \multicolumn{1}{c}{\scriptsize $\uparrow$} \\
\midrule
\multirow{4}{*}{\rotatebox{90}{\tiny Architecture}} & \textsc{RelAgent} & 78.34 & 85.23 & 68.36 & 67.79 & 78.20 & 87.25 & 71.86 & 70.78 & 82.84 & 90.41 & 88.42  & 71.07 & 78.38 \\
& \quad w/o feedback & 58.43          & 70.31          & 53.22          & 50.27          & 72.06          & 87.16          & 66.24 & 66.48          & 81.80          & 65.96          & 60.06          & 70.37 & 66.86\\
& \quad w/o eval workspace & 64.94 & 79.83 & 68.76 & 70.79 & 75.98 & 84.55 & 65.23 & 70.21 & 82.24 & 88.86 & 87.67 & 70.37 & 75.78 \\
& \quad w/o semantic & 65.44          & 77.70          & 67.09          & 67.42          & 73.39          & 81.23          & 62.20          & 70.23            & 82.32          & 89.45          & 85.70            & 71.07 & 74.44 \\
\midrule
\multirow{3}{*}{\rotatebox{90}{\tiny Predictor}} & \quad rule-only & 62.64 & 83.45 & 61.15 & 64.38 & 66.79 & 82.72 & 55.55 & 61.06 & 75.30 & 88.91 & 77.13 & 59.90 & 69.92 \\
& \quad catboost-only & 67.64 & 80.27 & 66.74 & 67.57 & 73.38 & 81.16 & 72.98 & 70.46 & 82.71 & 90.18 & 88.04 & 70.35 & 75.95 \\
& \quad lightgbm-only & 73.30 & 81.22 & 66.50 & 61.47 & 78.53 & 84.21 & 65.29 & 67.54 & 82.28 & 90.23 & 89.52 & 71.21 & 75.94 \\
\midrule
\multirow{2}{*}{\rotatebox{90}{\tiny LLM }} & \quad with Deepseek-V4 & 68.97 & 86.81 & 67.77 & 67.67 & 77.64 & 82.50 & 74.10 & 69.96 & 82.06 & 90.43 & 87.67 & 70.97 & 76.00 \\
& \quad with Qwen3-32B & 57.67 & 58.04 & 65.10 & 64.62 & 74.73 & 75.91 & 71.36 & 69.22 & 79.92 & 89.11 & 65.30 & 70.62 & 70.13  \\
\bottomrule
\end{tabular}
\end{table}

\subsection{Search setup}
For each task, we run five independent searches with up to 60 agent turns each.
The agent uses GPT-5.2 at temperature $1$ and up to $8{,}000$ output tokens per turn, with a $900$ second time-out per turn. 
Among the five searches, we select the one with the highest best-trial validation score as the final predictor. We refer to this process as cross-rollout model selection.
The agent selects among seven tree-based model families:
LightGBM with standard boosting (\texttt{gbdt}), DART (\texttt{dart}), and GOSS (\texttt{goss})~\citep{lightgbm};
XGBoost standard (\texttt{xgboost}) and XGBoost+DART (\texttt{xgb\_dart})~\citep{chen2016xgboost,gilad2015dart};
Random Forest (\texttt{rf}) via LightGBM; and CatBoost (\texttt{catboost})~\citep{prokhorenkova2017catboost}.
For each model choice the agent also tunes key hyperparameters: number of estimators $n_T$, learning rate $\eta$, and tree depth $d$.
For regression, the agent additionally selects the training objective ($L1$/MAE or $L2$/MSE) and whether to log-transform the target.
All SQL queries are executed via DuckDB~\citep{duckdb}. Prompts are reported in \Cref{app:prompts}.

\subsection{Selected programs}
Tables~\ref{tab:prog_cls} and~\ref{tab:prog_reg} report, for each task, the number of SQL feature queries in the selected program, the chosen downstream model, and the key hyperparameters of the best-trial configuration. Among all the model families, CatBoost is the most frequently selected model for classification tasks (14 of 21 tasks), followed by XGBoost-based variants. The number of SQL feature queries per task ranges from 3 to 19, with a median of 7.
For regression, the agent consistently selects L1/MAE objectives across all tasks, and applies a log-transformation of the target in 9 of 11 regression tasks, consistent with the skewed distributions common in LTV and activity-count targets.
\begin{table}[h]
\centering
\caption{Selected programs for classification tasks.
$n_T$: number of estimators; $\eta$: learning rate; $d$: tree depth;
\textbf{Val.}: best validation AUROC (\%) of the selected run;
\textbf{Test}: test AUROC (\%) of the selected run;
\textbf{Std}: standard deviation of test AUROC across 5 runs.}
\label{tab:prog_cls}
\setlength{\tabcolsep}{4pt}
\footnotesize
\begin{tabular}{llrlrrr|rrr}
\toprule
\textbf{Benchmark} & \textbf{Task} & \textbf{\#SQL} & \textbf{Model} & $n_T$ & $\eta$ & $d$ & \textbf{Val.} & \textbf{Test} & \textbf{Std} \\
\midrule
\multirow{12}{*}{\texttt{RelBenchV1}}
 & \texttt{f1}/\texttt{driver-dnf}           & 7  & \texttt{gbdt}     & 500 & 0.030 & 5 & 66.44 & 78.34 & 3.72 \\
 & \texttt{f1}/\texttt{driver-top3}          & 10 & \texttt{catboost} & 500 & 0.030 & 6 & 74.74 & 85.23 & 6.89 \\
 & \texttt{avito}/\texttt{user-clicks}       & 9  & \texttt{catboost} & 500 & 0.050 & 8 & 67.11 & 68.36 & 1.69 \\
 & \texttt{avito}/\texttt{user-visits}       & 11 & \texttt{catboost} & 500 & 0.032 & 5 & 71.29 & 67.79 & 0.12 \\
 & \texttt{event}/\texttt{user-repeat}       & 7  & \texttt{xgboost}  & 500 & 0.050 & 5 & 78.21 & 78.20 & 2.94 \\
 & \texttt{event}/\texttt{user-ignore}       & 5  & \texttt{catboost} & 400 & 0.050 & 6 & 76.66 & 87.25 & 1.30 \\
 & \texttt{trial}/\texttt{study-outcome}     & 3  & \texttt{catboost} & 500 & 0.012 & 5 & 66.19 & 71.86 & 2.18 \\
 & \texttt{amazon}/\texttt{user-churn}       & 11 & \texttt{catboost} & 500 & 0.060 & 8 & 70.65 & 70.78 & 0.51 \\
 & \texttt{amazon}/\texttt{item-churn}       & 9  & \texttt{catboost} & 500 & 0.060 & 9 & 82.39 & 82.84 & 0.23 \\
 & \texttt{stack}/\texttt{user-engagement}   & 7  & \texttt{xgb\_dart}& 500 & 0.050 & 6 & 90.33 & 90.41 & 0.19 \\
 & \texttt{stack}/\texttt{user-badge}        & 13 & \texttt{xgboost}  & 500 & 0.050 & 7 & 89.65 & 88.42 & 1.27 \\
 & \texttt{hm}/\texttt{user-churn}           & 19 & \texttt{catboost} & 500 & 0.050 & 9 & 71.50 & 71.07 & 0.27 \\
\midrule
\multirow{4}{*}{\texttt{RelBenchV2}}
 & \texttt{ratebeer}/\texttt{beer-churn}     & 5  & \texttt{catboost} & 500 & 0.060 & 8 & 90.27 & 84.70 & 0.93 \\
 & \texttt{ratebeer}/\texttt{user-churn}     & 3  & \texttt{xgb\_dart}& 500 & 0.050 & 9 & 98.90 & 98.63 & 0.83 \\
 & \texttt{ratebeer}/\texttt{brewer-dormant} & 15 & \texttt{catboost} & 500 & 0.050 & 8 & 84.80 & 83.33 & 0.33 \\
 & \texttt{arxiv}/\texttt{paper-citation}    & 8  & \texttt{catboost} & 500 & 0.050 & 6 & 82.47 & 82.62 & 0.14 \\
\midrule
\multirow{5}{*}{\texttt{4DBInfer}}
 & \texttt{amazon}/\texttt{user-churn}        & 13 & \texttt{catboost} & 500 & 0.050 & 8 & 79.90 & 79.44 & 0.25 \\
 & \texttt{outbrain}/\texttt{ad-ctr}          & 4  & \texttt{catboost} & 500 & 0.040 & 4 & 63.89 & 62.63 & 4.42 \\
 & \texttt{retailrocket}/\texttt{item-cvr}    & 7  & \texttt{catboost} & 500 & 0.060 & 8 & 88.65 & 86.35 & 0.73 \\
 & \texttt{stackexchange}/\texttt{post-upvote}& 13 & \texttt{xgboost}  & 500 & 0.050 & 7 & 96.24 & 89.45 & 3.31 \\
 & \texttt{stackexchange}/\texttt{user-churn} & 12 & \texttt{dart}     & 500 & 0.050 & 7 & 87.16 & 89.03 & 0.21 \\
\bottomrule
\end{tabular}
\end{table}

\begin{table}[h]
\centering
\caption{Selected programs for regression tasks.
All selected programs use an L1 (MAE) objective.
The ``Log'' column indicates whether the agent applied a log-transformation to the regression target;
MAE is reported in the original space after inverting the transformation.
\textbf{Val.}: best validation MAE (\%) of the selected run;
\textbf{Test}: test MAE of the selected run;
\textbf{Std}: standard deviation of test MAE across 5 runs.}
\label{tab:prog_reg}
\setlength{\tabcolsep}{2pt}
\footnotesize
\begin{tabular}{llcrrrrcc|rrr}
\toprule
\textbf{Benchmark} & \textbf{Task} & \textbf{\#SQL} & \textbf{Model} & $n_T$ & $\eta$ & $d$ & \textbf{Obj.} & \textbf{Log} & \textbf{Val.} & \textbf{Test} & \textbf{Std} \\
\midrule
\multirow{9}{*}{\texttt{RelBenchV1}}
 & \texttt{f1}/\texttt{driver-position}       & 10 & \texttt{goss}     & 500 & 0.050 & 6  & L1 & --- & 3.598 & 4.019 & 0.153 \\
 & \texttt{avito}/\texttt{ad-ctr}             & 5  & \texttt{gbdt}     & 500 & 0.050 & 7  & L1 & $\checkmark$ & 0.030 & 0.033 & 0.000 \\
 & \texttt{event}/\texttt{user-attendance}    & 7  & \texttt{dart}     & 500 & 0.040 & 4  & L1 & $\checkmark$ & 0.238 & 0.241 & 0.009 \\
 & \texttt{trial}/\texttt{study-adverse}      & 7  & \texttt{xgboost}  & 500 & 0.070 & 7  & L1 & $\checkmark$ & 37.439 & 37.194 & 3.553 \\
 & \texttt{trial}/\texttt{site-success}       & 5  & \texttt{xgboost}  & 500 & 0.050 & 6  & L1 & --- & 0.400 & 0.386 & 0.018 \\
 & \texttt{amazon}/\texttt{user-ltv}          & 7  & \texttt{xgboost}  & 500 & 0.030 & 7  & L1 & $\checkmark$ & 11.771 & 13.949 & 0.060 \\
 & \texttt{amazon}/\texttt{item-ltv}          & 5  & \texttt{xgboost}  & 500 & 0.050 & 8  & L1 & $\checkmark$ & 36.370 & 41.765 & 0.332 \\
 & \texttt{stack}/\texttt{post-votes}         & 7  & \texttt{xgboost}  & 500 & 0.040 & 8  & L1 & $\checkmark$ & 0.058 & 0.064 & 0.001 \\
 & \texttt{hm}/\texttt{item-sales}            & 7  & \texttt{xgb\_dart}& 500 & 0.050 & 8  & L1 & $\checkmark$ & 0.043 & 0.035 & 0.001 \\
\midrule
\multirow{2}{*}{\texttt{RelBenchV2}}
 & \texttt{ratebeer}/\texttt{user-count}         & 6  & \texttt{xgboost} & 500 & 0.040 & 8  & L1 & $\checkmark$ & 4.367 & 6.021 & 0.107 \\
 & \texttt{arxiv}/\texttt{author-publication}    & 7  & \texttt{goss}    & 500 & 0.025 & 10 & L1 & $\checkmark$ & 0.420 & 0.462 & 0.016 \\
\bottomrule
\end{tabular}
\end{table}

\paragraph{Variance among single searches.} Tables~\ref{tab:prog_cls} and~\ref{tab:prog_reg} also report the standard deviation of test scores across the 5 searches that were performed for the cross-rollout model selection.
On most tasks, the standard deviation is below 2 AUROC points (classification) or small relative to the score magnitude (regression), indicating that individual runs are broadly representative of overall performance.
Tasks with higher variance (e.g., \texttt{f1}/\texttt{driver-top3} $\sigma{=}6.89$, \texttt{trial}/\texttt{study-outcome} $\sigma{=}2.18$) tend to be smaller datasets where randomness in program search has a larger effect.

\paragraph{Variance among cross-rollout model selections.}
For entity classification tasks in RelBenchV2, we additionally report the mean and standard deviation of test AUROC among 3 independent cross-rollout model selections, each consisting of 5 searches (Table~\ref{tab:variance_cross_rollout}). Due to resource constraints, we omit other tasks and datasets. Across all four tasks, the standard deviation remains below $0.25$ AUROC points, indicating that this best-of-$K$ selection procedure yields stable final performance with low sensitivity to the randomness of individual rollouts.

\paragraph{Validation-test distribution shift.} On some small datasets (e.g., \texttt{f1}/\texttt{driver-top3} and \texttt{trial}/\texttt{study-outcome}), \textbf{Val.} exceeds \textbf{Test}, suggesting mild overfitting to the validation split; however, these gaps are small and primarily occur in low-sample regimes. Most datasets show closely matched validation and test performance, indicating that validation performance is generally a reliable criterion for model selection rather than a source of systematic overfitting.

\paragraph{Relational invariance.}
Relational invariance, i.e., invariance to permutations of row identifiers and table orderings, is induced by the selected SQL feature queries: if no query depends on row identifiers, the resulting predictions are invariant as well. In our experiments, queries that directly access row identifiers were not explicitly forbidden. We therefore inspected all final selected feature programs across all tasks and found that none accesses a row identifier of any table.  We did, however, find instances in which a query conditions on a categorical identifier, e.g., \texttt{sales\_channel\_id} in \texttt{hm}/\texttt{user-churn}. This does not violate relational invariance: such identifiers are fixed categorical encodings with semantic meaning (e.g., online vs.\ offline), not row identifiers, and are thus invariant to row permutations.

\begin{table}[t]
\footnotesize
  \centering
  \caption{Mean and standard deviations of test AUROC (\%) among 3 cross-rollout model selections with each 5 searches for entity classification tasks in RelBenchV2.}
  \begin{tabular}{l|rrrr}
    \toprule
     \textbf{Task} & \texttt{arxiv/citation} & \texttt{ratebeer/beer }& \texttt{ratebeer/dormant} & \texttt{ratebeer/user} \\
    \midrule
    \textbf{Mean}     & $82.59$ & $84.55$ & $83.01$ & $98.54$ \\
    \textbf{Std.}  & $0.031$ & $0.135$ & $0.239$ & $0.072$ \\
    \bottomrule
  \end{tabular}
  \label{tab:variance_cross_rollout}
\end{table}

\subsection{SQL query analysis}
To better understand the programs discovered by \textsc{RelAgent}, we analyze the selected program for each task (Tables~\ref{tab:prog_cls}--\ref{tab:prog_reg}). Across 32 selected programs, the agent produces 264 feature queries, with a median of 7 queries per program and a range of 3--19. Note that no two selected queries are identical, suggesting that the agent synthesizes task-specific SQL programs rather than reusing fixed templates. \Cref{tab:sql_analysis} summarizes the structural properties of these queries.

\begin{table}[h]
\centering
\caption{Structural properties of the 264 feature queries across all 32 selected programs (one per task).
All percentage counts refer to the fraction of queries (out of 264) that exhibit each property.}
\label{tab:sql_analysis}
\setlength{\tabcolsep}{5pt}
\footnotesize
\begin{tabular}{lc}
\toprule
\textbf{Property} & \textbf{\# queries (\%)} \\
\midrule
\multicolumn{2}{l}{\textit{Query structure}} \\
\quad Queries per program (median / range) & 7 \ / \ 3--19 \\
\quad Joins per query (mean / median) & 1.84 \ / \ 1 \\
\quad 0 joins (direct scan) & \phantom{0}18 \ (7\%) \\
\quad 1 join & 131 \ (50\%) \\
\quad 2 joins & \phantom{0}58 \ (22\%) \\
\quad 3 joins & \phantom{0}24 \ (9\%) \\
\quad 4+ joins & \phantom{0}33 \ (13\%) \\
\quad Using CTEs (\texttt{WITH} clauses) & 186 \ (70\%) \\
\midrule
\multicolumn{2}{l}{\textit{Temporal patterns}} \\
\quad Contains interval-based temporal filter & 193 \ (73\%) \\
\quad Contains recency feature (days since last event) & \phantom{0}94 \ (36\%) \\
\quad Contains 7-day window & \phantom{0}40 \ (15\%) \\
\quad Contains 30-day window & \phantom{0}78 \ (30\%) \\
\quad Contains 90-day window & \phantom{0}70 \ (27\%) \\
\quad Contains 365-day window & \phantom{0}57 \ (22\%) \\
\quad Contains 180-day window & \phantom{0}40 \ (15\%) \\
\midrule
\multicolumn{2}{l}{\textit{Aggregate functions}} \\
\quad \texttt{COUNT} & 163 \ (62\%) \\
\quad \texttt{CASE}/\texttt{IF} conditional & 155 \ (59\%) \\
\quad \texttt{SUM} & 104 \ (39\%) \\
\quad \texttt{MAX}  & 83  \ (31\%) \\
\quad \texttt{AVG} & 80 \ (30\%) \\
\quad Division ratio features & \phantom{0}25 \ (9\%) \\
\bottomrule
\end{tabular}
\end{table}

\textbf{Join depth and structure.}
Half of all queries join exactly one additional table; 7\% are direct scans with no join; 44\% join two or more tables (maximum 9).
Common Table Expressions (CTEs, written as \texttt{WITH name AS (\dots)}) appear in 70\% of queries, reflecting a preference for staged, named intermediate results over deeply nested subqueries.

\textbf{Temporal structure.}
Every query references \texttt{eval\_table}, which exposes the target row and its prediction timestamp. This enforces row-wise alignment by construction. For timestamped event tables, temporal non-leakage is achieved through explicit predicates comparing event timestamps against the corresponding \texttt{eval\_table} cutoff; 73\% of queries contain interval-based temporal filters, while the remaining queries are mostly static/profile features or non-windowed aggregates.
Surprisingly, the agent independently arrives at the same canonical aggregation horizons, e.g., 30, 90, 365, 180, 7, 14, and 60 days, that human analysts use for churn, LTV, and activity modelling, with no instruction to do so.

\textbf{Aggregate repertoire.}
\texttt{COUNT} is the most common aggregate (62\%), followed by conditional logic via \texttt{CASE}/\texttt{IF} expressions (59\%), \texttt{SUM} (39\%), and \texttt{MAX}/\texttt{AVG} (31\%/30\%).
Division-based ratio features (e.g., recent activity rate over lifetime count) appear in 9\% of queries.

\section{Prompts}
\label[appendix]{app:prompts}

We list the prompts sent to the agent at each stage of a run. Curly-brace tokens (\texttt{\{entity\_col\}}, \texttt{\{target\_col\}}, etc.) are filled in per task at runtime. The system prompt is sent once at the start of each run. Follow-up prompts are injected after each agent turn until the turn budget is exhausted, at which point the wrap-up prompt is sent instead.

\subsection{Tool descriptions}

The agent has access to the following tools. Descriptions are passed
to the LLM as part of the function-calling schema.

\begin{tcolorbox}[promptbox, title=\texttt{execute\_query}]
\begin{verbatim}
Execute a SQL query and return results.

Args:
    query (str): The SQL query to execute.

Returns:
    For SELECT queries: list of dicts (column -> value per row).
    For non-SELECT operations: status dict with 'status' and 'message'.
    For errors: error message string starting with "Error:".

Note: SELECT queries without a LIMIT clause are automatically
capped at 200 rows.
\end{verbatim}
\end{tcolorbox}

\begin{tcolorbox}[promptbox, title=\texttt{get\_table\_info}]
\begin{verbatim}
Get comprehensive information about table(s) in the database.

Returns schema, primary keys, foreign keys, and row counts.
If table_name is provided, returns info for that table;
otherwise returns info for all tables.

Args:
    table_name (str, optional): Name of a specific table.
        If None, returns info for all tables.
\end{verbatim}
\end{tcolorbox}

\begin{tcolorbox}[promptbox, title=\texttt{validate\_program} ]
\begin{verbatim}
Test a predictive program using a wrapped model on a data split.

Args:
    feature_queries_json: JSON string of feature queries.
        Format: [{"name": "query_name", "sql": "SELECT ..."}]
        Each query must be anchored on eval_table, return row_id,
        and produce feature columns for each target row. eval_table
        contains the split rows currently being scored, including
        row_id, {entity_col}, and {timestamp_col}. For timestamped
        source tables, feature queries must filter source records to
        occur before eval_table.{timestamp_col}.
        `train_table` is available and contains labeled training rows.
    model_choice: One of 7 learners: "gbdt", "rf", "dart", "goss",
        "xgboost", "xgb_dart", "catboost".
        - gbdt:     LightGBM standard gradient boosting
        - rf:       Random Forest via LightGBM
        - dart:     LightGBM with DART dropout regularization
        - goss:     LightGBM with gradient-based subsampling
        - xgboost:  XGBoost (second-order gradients)
        - xgb_dart: XGBoost with DART dropout
        - catboost: CatBoost (ordered boosting)
    model_config_json: JSON dict of hyperparameters (optional).
        Common keys: n_estimators, learning_rate, max_depth,
                     subsample, colsample_bytree.
        LightGBM variants also accept: min_child_samples,
                     lambda_l1, lambda_l2.
        XGBoost variants also accept: min_child_weight,
                     reg_alpha, reg_lambda.
        CatBoost also accepts: l2_leaf_reg.
        Out-of-bounds values are clamped. Unknown keys are ignored.

Returns:
    Formatted string with metrics, diagnostics, and trial history.
\end{verbatim}
\end{tcolorbox}

\begin{tcolorbox}[promptbox, title=\texttt{get\_trial\_history}]
\begin{verbatim}
Get a summary of all previous validation trials.

Returns:
    Formatted string showing trial history with scores and
    approach summaries.
\end{verbatim}
\end{tcolorbox}

\begin{tcolorbox}[promptbox, title=\texttt{query\_eval\_workspace}]
\begin{verbatim}
Query the evaluation workspace to analyse trial results.

The workspace is a DuckDB database with the following tables:

  trials
    trial_id TEXT, trial_name TEXT, parent_trial_id TEXT,
    created_at TIMESTAMPTZ, split TEXT, model_choice TEXT,
    resolved_model_config TEXT, feature_query_hash TEXT,
    feature_block_names TEXT, primary_metric TEXT,
    primary_score DOUBLE, metrics_json TEXT, notes TEXT


  eval_predictions
    trial_id TEXT, row_id INTEGER, entity_id TEXT, label TEXT,
    score DOUBLE, predicted_class TEXT, split TEXT,
    eval_cutoff TIMESTAMPTZ

row_id is stable across trials for the same split, so you can
join two trials on row_id to compare predictions on the same examples.

IMPORTANT: These tables are analysis artifacts only. Do NOT use
them as feature sources in your SQL feature queries.

Args:
    sql: A SELECT query to run against the workspace.

Returns:
    Query results (up to 500 rows) or an error message.
\end{verbatim}
\end{tcolorbox}

\subsection{System prompt}

\begin{tcolorbox}[promptbox, title={System Prompt: Classification}]
\begin{verbatim}
You are a data scientist building a predictive pipeline for a
ENTITY CLASSIFICATION task.

GOAL: Find a set of SQL feature queries + a model choice that
accurately predict the class label ({target_col}) for ALL entities
in the validation set.

You have the following tools:
1. SQL tools (execute_query, get_table_info, etc.)
   -- explore and query the database
2. validate_program(feature_queries_json, model_choice, model_config_json)
   -- train and evaluate your feature pipeline on the validation split
3. get_trial_history()
   -- see what you've already tried and their scores
4. query_eval_workspace(sql)
   -- analyze the evaluation workspace (trials, eval_predictions)

Rules:
- Use SQL tools to explore the database. Do NOT guess table/column names.
- Start by running SHOW TABLES and PRAGMA table_info('table') to
  understand the schema.
- train_table contains labeled training examples. Use it to learn patterns.
- eval_table contains the current rows being scored (val/test input keys).
- Always anchor feature SQL on eval_table (not train_table) -- eval_table
  is swapped with the correct split at each call.
- Write SQL queries that extract features PER TARGET ROW.
- eval_table contains one row per target example, including row_id,
  {entity_col}, and {timestamp_col}.
- Each feature query must be anchored on eval_table and return row_id.
- For timestamped source tables, include temporal filters of the form
  source_time < eval_table.{timestamp_col}.
- Do not group only by {entity_col}: the same entity may appear at
  multiple prediction timestamps. Group by row_id, and merge features
  back by row_id.
- Call validate_program() with your SQL feature blocks, model_choice,
  and optional model_config_json to run on the validation set.
- Iterate based on the metrics, diagnostics, and error examples returned.

- For BINARY classification, the model outputs probability scores (0-1).
- Think about features like: entity frequency, recency, aggregates from
  related tables.
- The validation tool returns metrics (AUROC, F1, etc.)
  + diagnostics.

=== WRAPPED MODEL MODE ===

Your validate_program() tool accepts:
  - feature_queries_json: SQL feature queries (same as before)
  - model_choice: one of the 7 learners below
  - model_config_json: optional JSON dict of hyperparameters

Available models:
  1. "gbdt"     -- Standard Gradient Boosted Trees. Fast, strong default.
                   Config: n_estimators (50-500), learning_rate (0.01-0.3),
                           max_depth (2-10), min_child_samples (1-100),
                           subsample (0.5-1.0), colsample_bytree (0.5-1.0)
                   Regularization: lambda_l1 (0.0-10.0), lambda_l2 (0.0-10.0)
  2. "rf"       -- Random Forest (bagging; less sensitive to learning rate).
                   Config: same keys as gbdt
  3. "dart"     -- DART Boosting (dropout regularization).
                   Config: same keys as gbdt
  4. "goss"     -- GOSS (gradient-based subsampling; fast on large datasets).
                   Config: same keys as gbdt
  5. "xgboost"  -- XGBoost (second-order gradients; different regularization).
                   Config: n_estimators (50-500), learning_rate (0.01-0.3),
                           max_depth (2-10), min_child_weight (1-100),
                           subsample (0.5-1.0), colsample_bytree (0.5-1.0)
                   Regularization: reg_alpha (0.0-10.0), reg_lambda (0.0-10.0)
  6. "xgb_dart" -- XGBoost + DART dropout.
                   Config: same keys as xgboost
  7. "catboost" -- CatBoost (ordered boosting; robust on heterogeneous features).
                   Config: n_estimators (50-500), learning_rate (0.01-0.3),
                           max_depth (2-10), l2_leaf_reg (0.1-10.0)

Categorical features (all 7 learners):
  Add "categorical_features" inside model_config_json as a list of
  "<query_name>__<col>" names to treat columns natively as categorical.
  High-cardinality columns (>~4000 levels) should be bucketed in SQL first.

Constraints:
  - Do NOT write free-form training code, custom objectives, or ensembles.
  - Do NOT perform hyperparameter search loops -- one config per call.
  - Out-of-bounds config values are clamped automatically.
  - Omitted config fields use strong defaults.
  - The environment handles train/val splitting, fitting, and evaluation.

The tool returns: metrics, resolved model config, row counts after merges,
missingness rates, warnings, and best/worst prediction examples.
Use this feedback to iterate on your SQL features.

=== EVALUATION WORKSPACE ===
After each validate_program() call, the full evaluation output is
persisted to a queryable workspace.
Use query_eval_workspace(sql) to analyse results.

Workspace tables:

  trials
    trial_id TEXT, trial_name TEXT, parent_trial_id TEXT,
    created_at TIMESTAMPTZ, split TEXT, model_choice TEXT,
    resolved_model_config TEXT, feature_query_hash TEXT,
    feature_block_names TEXT, primary_metric TEXT,
    primary_score DOUBLE, metrics_json TEXT, notes TEXT

  eval_predictions
    trial_id TEXT, row_id INTEGER, entity_id TEXT, label TEXT,
    score DOUBLE, predicted_class TEXT, split TEXT,
    eval_cutoff TIMESTAMPTZ

row_id is positionally stable across trials for the same split, so
you can join two trials on row_id to compare predictions on the same
examples.

IMPORTANT: Always run PRAGMA table_info('tablename') to verify exact
column names. Column names differ across datasets -- never assume a
column like 'id', 'type', 'date', or 'count' exists without checking.
\end{verbatim}
\end{tcolorbox}

The regression system prompt is identical to the classification one
above, except for the goal statement, task-type bullet points, and the
model menu (each learner gains a regression objective key:
\begin{itemize}
\item \texttt{regression\_l1}/\texttt{regression\_l2}/\texttt{huber} for LightGBM variants,
\item \texttt{reg:absoluteerror}/\texttt{reg:squarederror} for XGBoost variants).
\end{itemize}

\begin{tcolorbox}[promptbox]
\begin{verbatim}
You are a data scientist building a predictive pipeline for a
ENTITY REGRESSION task.

GOAL: Find a set of SQL feature queries + a model choice that
accurately predict the numerical value ({target_col}) for ALL
entities in the validation set.

[... tool inventory and Rules identical to classification ...]

- The model outputs a numeric value for each entity.
- The validation tool returns metrics (MAE)
  + diagnostics.
- Think about: historical averages, trends, aggregates from related
  tables.
- Primary objective for this run is MAE: minimize absolute error.


=== WRAPPED MODEL MODE ===

Your validate_program() tool accepts:
  - feature_queries_json: SQL feature queries (same as before)
  - model_choice: one of the 7 learners below
  - model_config_json: optional JSON dict of hyperparameters

Available models:
  1. "gbdt"     -- Standard Gradient Boosted Trees. Fast, strong default.
                   Config: n_estimators (50-500), learning_rate (0.01-0.3),
                           max_depth (2-10), min_child_samples (1-100),
                           subsample (0.5-1.0), colsample_bytree (0.5-1.0)
                   objective: "regression_l1" (MAE), "regression_l2" (MSE),
                              "huber"
                   Default: "regression_l1" -- directly minimises eval metric.
  2. "rf"       -- Random Forest (bagging; less sensitive to learning rate).
                   Config: same keys as gbdt (including objective)
  3. "dart"     -- DART Boosting (dropout regularization).
                   Config: same keys as gbdt (including objective)
  4. "goss"     -- GOSS (gradient-based subsampling; fast on large datasets).
                   Config: same keys as gbdt (including objective)
  5. "xgboost"  -- XGBoost (second-order gradients).
                   Config: n_estimators (50-500), learning_rate (0.01-0.3),
                           max_depth (2-10), min_child_weight (1-100),
                           subsample (0.5-1.0), colsample_bytree (0.5-1.0)
                   objective: "reg:absoluteerror" (MAE),
                              "reg:squarederror" (MSE),
                              "reg:pseudohubererror" (Huber)
  6. "xgb_dart" -- XGBoost + DART dropout.
                   Config: same keys as xgboost (including objective)
  7. "catboost" -- CatBoost (ordered boosting).
                   Config: n_estimators (50-500), learning_rate (0.01-0.3),
                           max_depth (2-10), l2_leaf_reg (0.1-10.0)

  For skewed targets: add "log_transform_target": true to
  model_config_json. The harness fits on log1p(y) and reports MAE
  back in the original scale.

Categorical features (all 7 learners):
  Add "categorical_features" inside model_config_json as a list of
  "<query_name>__<col>" names to treat columns natively as categorical.

Constraints:
  - Do NOT write free-form training code, custom objectives, or ensembles.
  - Do NOT perform hyperparameter search loops -- one config per call.
  - Out-of-bounds config values are clamped automatically.
  - The environment handles train/val splitting, fitting, and evaluation.

[... feedback description and Evaluation Workspace identical to classification ...]
\end{verbatim}
\end{tcolorbox}

\subsection{Execution Prompts}

\begin{tcolorbox}[promptbox,title={Execute Prompt (First User Message)}]
\begin{verbatim}
Task: {task_type}
Task Description: {task_description}
Dataset: {dataset_name}

Entity column: {entity_col}
Timestamp column: {timestamp_col}
Target column: {target_col}
Validation set size: {n_val} target rows

WRAPPED MODEL MODE: You propose SQL features + a model choice.
The environment trains and evaluates.

Steps:
1. Run SHOW TABLES to see available tables
2. Run PRAGMA table_info('train_table') and inspect other tables
3. Explore data distributions (SELECT COUNT(*), sample rows, etc.)
4. Design SQL feature queries that extract per-target-row signals
   anchored on eval_table and keyed by row_id
5. Call validate_program() with your features, model_choice, and
   optional config

Example call:
  validate_program(
    feature_queries_json='[{"name": "basic_stats",
                            "sql": "SELECT ..."}]',
    model_choice="gbdt",
    model_config_json='{}'
  )

Start with gbdt and simple features. Iterate by improving SQL
features based on diagnostics.

You have query_eval_workspace(sql): after each validate_program(),
analyze trials and eval_predictions as described in the system prompt.
\end{verbatim}
\end{tcolorbox}

\begin{tcolorbox}[promptbox,title={Follow-up Prompt (Intermediate Turns)}]
\begin{verbatim}
Continue improving your pipeline.
- If you haven't validated yet, call validate_program() now.
- Full evaluation results are available in the workspace after each trial.
  Use query_eval_workspace() for error analysis.
- Call get_trial_history() to see all past attempts.
- Use what you find in the workspace to guide your next SQL feature
  improvements.
\end{verbatim}
\end{tcolorbox}

\begin{tcolorbox}[promptbox,title={Wrapup Prompt (Final Turn)}]
\begin{verbatim}
This is your last turn. If you haven't submitted a validation yet, do so now. 
Call validate_program() with your best SQL queries, model choice, and model configuration. 
If you already have results, call get_trial_history() to confirm your best score.
\end{verbatim}
\end{tcolorbox}

\subsection{Task Descriptions}
\label{app:task-descriptions}

Tables~\ref{tab:task-desc-v1}--\ref{tab:task-desc-4db} list the
\texttt{\{task\_description\}} string injected into the execute prompt for each task.

\begin{table}[t]
\centering
\small
\caption{Task descriptions of RelBench V1 (12 classification, 9 regression).}
\label{tab:task-desc-v1}
\begin{tabular}{@{}lllp{7.2cm}@{}}
\toprule
\textbf{Dataset} & \textbf{Task} & \textbf{Type} & \textbf{Description} \\
\midrule
\multirow{4}{*}{\texttt{rel-amazon}}
  & \texttt{user-churn}  & Cls & Predict 1 if the customer does not review any product in the next 3 months, and 0 otherwise. \\
  & \texttt{item-churn}  & Cls & Predict 1 if the product does not receive any reviews in the next 3 months. \\
  & \texttt{user-ltv}    & Reg & Predict the \$ value of the total products a user buys and reviews in the next 3 months. \\
  & \texttt{item-ltv}    & Reg & Predict the \$ value of the total purchases and reviews a product receives in the next 3 months. \\
\midrule
\multirow{3}{*}{\texttt{rel-stack}}
  & \texttt{user-engagement} & Cls & Predict if a user will make any votes, posts, or comments in the next 3 months. \\
  & \texttt{user-badge}      & Cls & Predict if a user will receive a new badge in the next 3 months. \\
  & \texttt{post-votes}      & Reg & Predict how many votes a user post will receive in the next 3 months. \\
\midrule
\multirow{3}{*}{\texttt{rel-trial}}
  & \texttt{study-outcome}  & Cls & Predict if the trial will achieve its primary outcome ($p < 0.05$). \\
  & \texttt{study-adverse}  & Reg & Predict the number of patients with severe adverse events or death. \\
  & \texttt{site-success}   & Reg & Predict the success rate of a trial site in the next 1 year. \\
\midrule
\multirow{3}{*}{\texttt{rel-f1}}
  & \texttt{driver-dnf}      & Cls & Predict if a driver will DNF in a race in the next 1 month. \\
  & \texttt{driver-top3}     & Cls & Predict if a driver will qualify in the top 3 for a race in the next 1 month. \\
  & \texttt{driver-position} & Reg & Predict the average finishing position of a driver across all races in the next 2 months. \\
\midrule
\multirow{2}{*}{\texttt{rel-hm}}
  & \texttt{user-churn}  & Cls & Predict whether a customer will have no transactions in the next week. \\
  & \texttt{item-sales}  & Reg & Predict the total sales (sum of prices) for an article in the next week. \\
\midrule
\multirow{3}{*}{\texttt{rel-event}}
  & \texttt{user-repeat}     & Cls & Predict if a user will attend an event in the next 7 days (given attendance in the last 14 days). \\
  & \texttt{user-ignore}     & Cls & Predict if a user will ignore more than 2 event invitations in the next 7 days. \\
  & \texttt{user-attendance} & Reg & Predict how many events a user will respond yes/maybe to in the next 7 days. \\
\midrule
\multirow{3}{*}{\texttt{rel-avito}}
  & \texttt{user-visits} & Cls & Predict whether a user will visit more than one ad in the next 4 days. \\
  & \texttt{user-clicks} & Cls & Predict whether a user will click on more than one ad in the next 4 days. \\
  & \texttt{ad-ctr}      & Reg & Given that an ad will be clicked in the next 4 days, predict its click-through rate. \\
\bottomrule
\end{tabular}
\end{table}

\begin{table}[t]
\centering
\small
\caption{Task descriptions of RelBench V2 (4 classification, 2 regression).}
\label{tab:task-desc-v2}
\begin{tabular}{@{}lllp{7.2cm}@{}}
\toprule
\textbf{Dataset} & \textbf{Task} & \textbf{Type} & \textbf{Description} \\
\midrule
\multirow{2}{*}{\texttt{rel-arxiv}}
  & \texttt{paper-citation}     & Cls & Predict whether a paper will receive at least one citation in the next 6 months. \\
  & \texttt{author-publication} & Reg & Predict how many papers an author will publish in the next 6 months. \\
\midrule
\multirow{4}{*}{\texttt{rel-ratebeer}}
  & \texttt{beer-churn}     & Cls & Predict whether a beer will receive no rating in the next 90 days. \\
  & \texttt{user-churn}     & Cls & Predict whether a user will give no beer rating in the next 90 days. \\
  & \texttt{brewer-dormant} & Cls & Predict whether a brewer will release zero new beers in the next 365 days. \\
  & \texttt{user-count}     & Reg & Predict the number of beer ratings a user will give in the next 90 days. \\
\bottomrule
\end{tabular}
\end{table}

\begin{table}[t]
\centering
\small
\caption{Task descriptions of 4DBInfer (5 classification).}
\label{tab:task-desc-4db}
\begin{tabular}{@{}lllp{7.2cm}@{}}
\toprule
\textbf{Dataset} & \textbf{Task} & \textbf{Type} & \textbf{Description} \\
\midrule
\texttt{amazon}        & \texttt{user-churn}   & Cls & Predict whether an Amazon reviewer will stop writing reviews in the next time window. \\
\midrule
\texttt{outbrain}      & \texttt{ad-ctr}       & Cls & Predict whether a user will click on a promoted content recommendation on Outbrain. \\
\midrule
\texttt{retailrocket}  & \texttt{item-cvr}     & Cls & Predict whether a user will convert (purchase) after viewing an item on RetailRocket. \\
\midrule
\multirow{2}{*}{\texttt{stackexchange}}
  & \texttt{post-upvote} & Cls & Predict whether a StackExchange post will receive an upvote. \\
  & \texttt{user-churn}  & Cls & Predict whether a StackExchange user will stop posting or participating. \\
\bottomrule
\end{tabular}
\end{table}

\section{Broader impacts}
\label[appendix]{app:broder_impacts}
\textsc{RelAgent} may reduce the manual effort required to build relational prediction systems and improve explainability by producing explicit SQL feature programs. However, relational databases may contain sensitive attributes or proxy variables, and an autonomous feature-search system may discover features that encode demographic, behavioral, or otherwise sensitive signals. The method should therefore be used with appropriate data governance, privacy controls, and fairness auditing, especially in high-stakes domains such as healthcare, finance, hiring, or public services.

\section{Limitations}
\label[appendix]{app:limitations}

\textsc{RelAgent} inherits both the strengths and weaknesses of SQL-based search. First, the search space grows combinatorially with the number of tables, join paths, time windows, filters, and aggregation operators, so the agent may miss useful deep or high-order relational features within a fixed turn budget. Second, although SQL can express many relational patterns, discovering the correct query can require brittle multi-step exploration. Third, the current implementation primarily exploits relational structures and statistics, and does not fully process long free-text fields or multimodal attributes. 
Finally, because the agent searches using validation feedback, care is required to prevent leakage and overfitting to the validation set.


\end{document}